\documentclass[10pt,journal,compsoc]{IEEEtran}

\usepackage{graphicx}
\usepackage{amsmath}
\usepackage{amssymb}
\usepackage{algorithmic}
\usepackage[ruled]{algorithm2e}
\usepackage{epsfig}
\usepackage{mathrsfs}
\usepackage{multirow}
\usepackage{booktabs}
\usepackage{mathtools}

\newcommand{\INPUT}{\item[\myinput]}
\newcommand{\myinput}{\textbf{Initialization:}}

\newcommand{\MYWHILE}{\item[\mywhile]}
\newcommand{\mywhile}{\textbf{repeat}}

\newcommand{\MYENDWHILE}{\item[\myendwhile]}
\newcommand{\myendwhile}{\textbf{until}}

\ifCLASSINFOpdf
  \DeclareGraphicsExtensions{.pdf,.jpeg,.png}
\else
  \DeclareGraphicsExtensions{.pdf}
\fi

\begin{document}
\title{Discriminatively Trained And-Or Graph Models for Object Shape Detection}

\author{Liang Lin, Xiaolong Wang, Wei Yang, and Jian-Huang Lai\thanks{This work was supported by the National Natural Science Foundation of
China (no. 61173082, no. 61173084), Guangdong Science and Technology Program (no. 2012B031500006), Guangdong Natural Science Foundation (no. S2013050014548), Special Project on Integration of Industry, Education and Research of Guangdong Province (no. 2012B091000101, no. 2012B091100148), and Fundamental Research Funds for the Central Universities (no. 13lgjc26).\protect\\Copyright (c) 2014 IEEE. Personal use of this material is permitted.
However, permission to use this material for any other purposes must be obtained from the IEEE by sending an email to pubs-permissions@ieee.org.}\IEEEcompsocitemizethanks{\IEEEcompsocthanksitem The authors are with Sun Yat-sen University, Guangzhou 510006, P. R. China, and L. Lin and J. H. Lai are with the SYSU-CMU Shunde International Joint Research Institute, Shunde, China. E-mail: linliang@ieee.org.
}

}

\markboth{IEEE TRANSACTIONS ON PATTERN ANALYSIS AND MACHINE INTELLIGENCE, 2014.}
{L. Lin \MakeLowercase{\textit{et al.}}: Discriminatively Trained And-Or Graph Models for Object Shape Detection}


%\vspace{-12mm}
\IEEEcompsoctitleabstractindextext{
\begin{abstract}

In this paper, we investigate a novel reconfigurable part-based model, namely And-Or graph model, to recognize object shapes in images. Our proposed model consists of four layers: leaf-nodes at the bottom are local classifiers for detecting contour fragments; or-nodes above the leaf-nodes function as the switches to activate their child leaf-nodes, making the model reconfigurable during inference; and-nodes in a higher layer capture holistic shape deformations; one root-node on the top, which is also an or-node, activates one of its child and-nodes to deal with large global variations (e.g. different poses and views).  We propose a novel structural optimization algorithm to discriminatively train the And-Or model from weakly annotated data.  This algorithm iteratively determines the model structures (e.g. the nodes and their layouts) along with the parameter learning. On several challenging datasets, our model demonstrates the effectiveness to perform robust shape-based object detection against background clutter and outperforms the other state-of-the-art approaches. We also release a new shape database with annotations, which includes more than $1500$ challenging shape instances, for recognition and detection. 

\end{abstract}



}

\maketitle

\IEEEdisplaynotcompsoctitleabstractindextext

\IEEEpeerreviewmaketitle

\begin{figure}[!htb]
\centering
\epsfig{figure=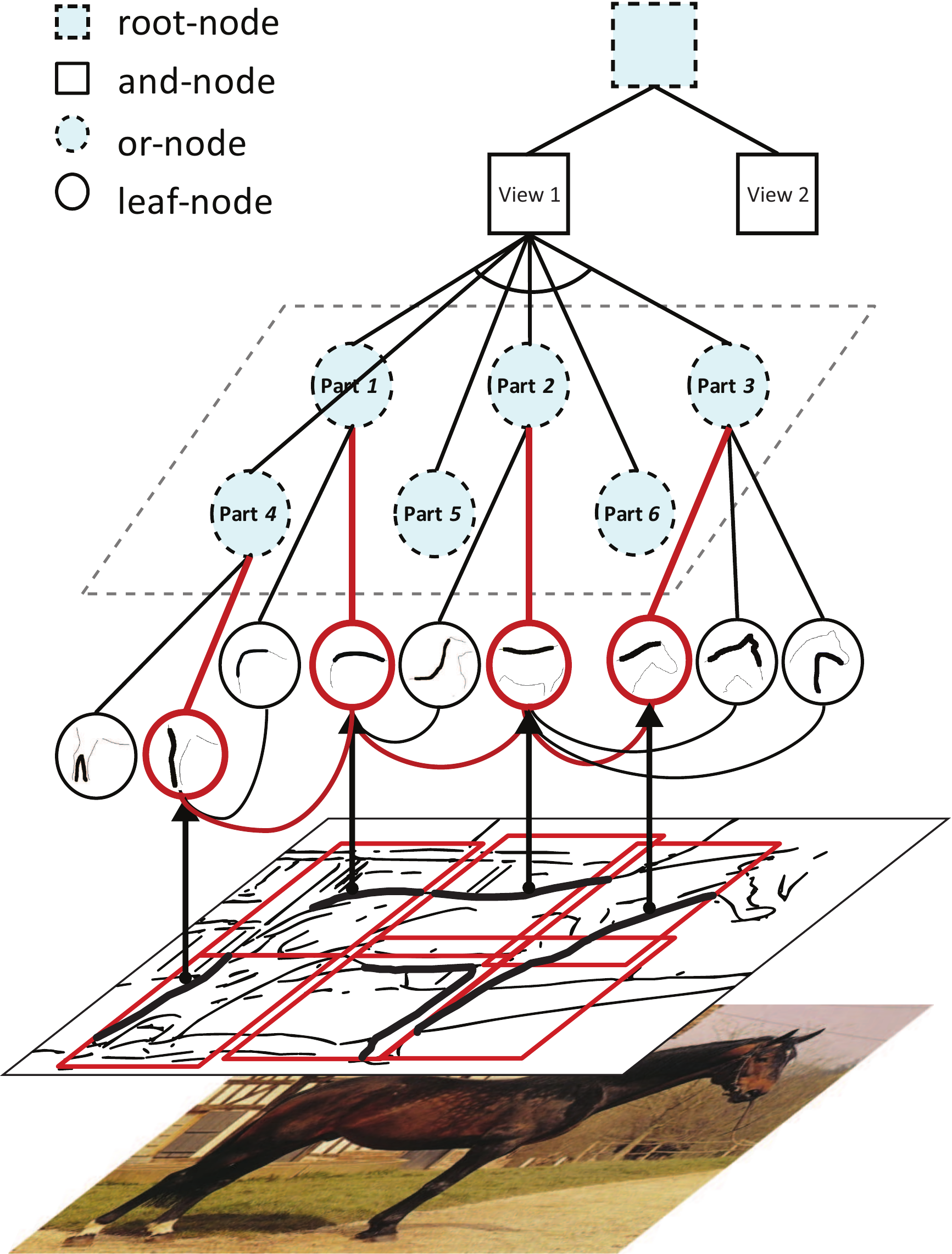,width=3.2in}
\caption{ An example of our And-Or graph model. It comprises four layers from bottom to top: the leaf-nodes (denoted by the solid circles) at the bottom for localizing local contour fragments, the or-nodes (denoted by the dashed blue circles) over the bottom specifying the activations of their child leaf-nodes, the and-nodes (denoted by the solid squares) encoding the holistic (view-based) variances, and the root-node (denoted by the dashed blue squares) on the top to switch the selection of its child and-nodes. The horizontal links incorporate contextual interactions among parts. Note that the leaf-nodes inherit the links that are defined between the layer of or-nodes. The nodes and links in red indicate the activation of leaf-nodes during the detection.  }\label{fig:AOG_example}
\end{figure}

\section{Introduction}\label{sec:introduction}

As psychophysics experiments suggested, humans can successfully identify objects in images using contour fragments alone~\cite{PsychologyContour}. In computer vision, recognizing object shapes from salient contours is an active research area. Several methods~\cite{ShiShapeCVPR2010,LateckiCVPR2011,ShapeParsing,PAS} have demonstrated that the contours (silhouettes) are robust against variations of illumination, color, and texture.  However, there are two long-standing difficulties in the current research.
\begin{itemize}
    \item Unreliable edge map extraction and contour tracing. Some key contours can be missing or connected to their background, making it difficult for accurately localizing shapes against surrounding clutter.
    \item Large variations within an object category, e.g. different object poses, views, occlusions, and deformations. Without using appearance or texture information, this challenge might be more serious, as shape contours are somewhat ambiguous and less discriminative.
\end{itemize}

Some recently proposed approaches addressed the two issues by learning hierarchical and compositional models, and achieved substantial progresses~\cite{ShottonPAMI08, ShiShapeCVPR2010,LinGraphMatch}. These models represent an object shape in terms of the parts (i.e. local contours) and the inter-part relations. However, their model structures (e.g. the number of parts and the ways of composition) are often fixed, consequently limiting the performances on complex scenarios.

In this work, we develop a novel reconfigurable part-based model in the form of an And-Or graph representation, which is discriminatively trained from weakly annotated training data (i.e. without annotating the object parts). Our model achieves superior performances on the task of detecting and localizing shapes from cluttered background, compared with other state-of-the-art methods.  Figure~\ref{fig:AOG_example} shows an example of our And-Or graph model.  The key component of our model is the ``switch variable'', referred to the or-node, which incorporates the compositional alternatives and makes the model reconfigurable. Specifically, the or-node specifies the way of compositions by activating the child nodes, to deal with the above-mentioned challenges in shape detection. Our And-Or graph model consists of four layers described as follows.

The \textbf{leaf-nodes} at the bottom represent a batch of local classifiers that detect the salient contour fragments of objects. Each leaf-node is defined within a  divided block, denoted by the red box in the bottom of Figure~\ref{fig:AOG_example}.  Given the edge map extracted from an image, a leaf-node takes the contours fallen into its block as the inputs. Once a long contour exceeds the block, it is automatically truncated. This is actually a partial matching scheme to handle the unreliable bottom-up edge tracing, i.e. to avoid object contours connecting to the background. Moreover, to capture the discriminability of contours, we design a new contour feature that combines the triangle-based descriptor~\cite{ShapeGroup} and the Shape Context descriptor~\cite{ShapeContext}.

The \textbf{or-nodes} defined as the switch variables that specify the activation of their child leaf-nodes, denoted by the dashed blue circles in Figure~\ref{fig:AOG_example}. During detection, each or-node activates one of its child lead-nodes and also selects the contour fragment detected by the activated leaf-node. The or-nodes thus represent the parts of an object shape, while the leaf-nodes capture all of the local variabilities. As Figure~\ref{fig:AOG_example} illustrates, our model can capture not only the local variations (e.g. part 2 of the example), but also the inconsistency caused by missing or broken edges (e.g. part 3 of the example).

The \textbf{collaborative edges} in our model impose the contextual information among shape contours, denoted by the horizontal links between the leaf-nodes in Figure~\ref{fig:AOG_example}.  Some of the existing compositional shape models ignore the contextual relations among contours, or simplify the relations by calculating the co-occurrence frequencies of neighbor contours~\cite{LinICCV07}. In contrast, we utilize informative spatial layout features to define the edges, motivated by the methods for contextualized object detection~\cite{DesaiIJCV,LateckiECCV2010}.

The \textbf{and-nodes}  aggregate the local shape contours that have been selected via the or-nodes.  Each and-node is defined as a potential function that captures the holistic shape deformations and distortions. Once the contour fragments are localized, The and-nodes further verify them as a whole to improve the discriminability of our model.

The \textbf{root-node} at the top functions as a switch to choose its child and-nodes, accounting for the large global variations (e.g. different views of shapes). It is defined exactly in the same way as the or-nodes. For example, two horses may appear diversely under different views, so that our model can adaptively activate different and-nodes for detecting them.

From the bottom to the top, our model is hierarchically constructed into an  ``And-Or-And-Or'' structure. Note that the leaf-nodes in our model can also be viewed as the and-nodes, as they are defined in the same way. This structure is very expressive and general to model object variations. The ``And'' symbol indicates the combination of sub-parts while the ``Or'' symbol indicates the switch between possible configurations.  We introduce the latent variables to make our model reconfigurable. In particular, the latent variables include the activation states of the or-nodes and the root-nodes, and the locations of contour fragments. The leaf-nodes and the and-nodes are defined as classification functions whose coefficients are treated as the observable model parameters. With the latent variables, the graph nodes and edges are explicitly mapped with the discriminative classification function of our model. Figure~\ref{fig:AoG_feature} provides an intuitive illustration of our And-Or graph model, which will be discussed later on. We regard our model as a general extension of the pictorial and deformable part-based models~\cite{Pictorial,HumanExp1,LatentSVM}, as it incorporates not only the hierarchical decompositions, but also the explicit structural alternatives.

The training of the And-Or graph model is another innovation of this work. The challenges lie in two aspects. First, multiple parameters in different layers need to be optimized along with the latent variables, and the objective function for optimization is non-convex, which cannot be solved directly with the traditional methods such as the support vector machines (SVMs).  Second, it is non-trivial to automatically discover the model structures in the model learning, as the training examples are not annotated into object parts. In the literature, learning And-Or graph models (or other reconfigurable models) usually relies on elaborative annotations or initializations~\cite{LeoAOG,AOGgrammar,LinGrammar}. To cope with these two problems, we propose a novel learning method, called Dynamical Structural Optimization (DSO), which is inspired by the recently proposed optimization methods~\cite{LatentSVM,CCCP,SVMICML2009}. This algorithm iteratively optimizes the model structures together with the multi-layer parameter learning, which includes three main steps. (i) Apply current model on the training examples while estimating the latent variables for each example. (ii) Discover new model structures. As the model structures are mapped with the discriminative function of our model (see Figure~\ref{fig:AoG_feature}), refactoring (rearranging) the feature vectors of training examples can lead to new structures. In brief, we perform clustering on the sub-feature-vectors corresponding to different nodes, and generate new structures according to the clustering results. For example, at one part of the shape, if the corresponding sub-feature-vectors are clustered into three groups, then we create three leaf-nodes accordingly to detect the local contours. (iii) Learn the model parameters with the newly generated structures.

Shape detection using the And-Or graph model is realized by searching over a image pyramid. We first accomplish two testing steps to generate several hypotheses of detection, and each hypothesis represents a configuration comprising detected contour fragments. (i) Local testing uses all leaf-nodes to detector contour fragments within the edge map. (ii) Binding testing imposes the collaborative edges among the contour fragments to further weigh the hypotheses. Afterwards, the and-nodes re-score each hypothesis by measuring the contour fragments as a whole. The root-node decides the final detection by selecting the most possible hypothesis.

%The key problem of training our And-Or graph model is automatic structure determination. We propose a novel learning algorithm, namely dynamic CCCP, extended from the concave-convex procedure (CCCP)~\cite{CCCP,SVMICML2009} by embedding the structural reconfiguration. It iterates to dynamically determine the production of leaf-nodes associated with the or-nodes, which is often simplified by manually fixing in previous methods~\cite{LeoCCCP,ShiShapeCVPR2010}. The other structure attributes (e.g., the layout of or-nodes and the activation of leaf-nodes) are implicitly inferred with the latent variables.

The remainder of this paper is organized as follows. Section \ref{sec:related-work} provides a brief review of related work. Then we present the model representations in Section \ref{sec:representation} and follow with a description of the inference procedure in Section \ref{sec:inference}. Section \ref{sec:learning} focuses on discussing the learning algorithm. The experimental results and comparisons are exhibited in Section \ref{sec:experiment}. Section \ref{sec:conclusion} concludes this paper.

\section{Related Work}\label{sec:related-work}

In this section, we review the extant techniques for shape (or contour) matching and shape model learning.

Many methods treat shape detection as a task of matching contours with certain distance measures, and they mostly utilized hand-drawn reference templates~\cite{ShapeContext,ShapeTree,ShiShapeECCV2008,BaiTransduction,ShapeGroup,PartialMatchingECCV2010,LateckiCVPR2011}. To handle diverse shape deformations and distortions, a number of robust shape (or contour) descriptors have been extensively discussed, such as Shape Context~\cite{ShapeContext}, Geodesic-Intensity Histogram~\cite{InnerDis}, Contour Flexibility~\cite{ConFlexibility}, and Local Angle~\cite{PartialMatchingECCV2010,ShapeGroup}. Based on these shape features,  several effective matching schemes~\cite{LinGraphMatch,TuShape,BaiTransduction} have been proposed to deal with the various challenges. For example, the inner-distance matching algorithm~\cite{InnerDis} was presented to handle the articulated shape deformations. Tu et al.~\cite{TuShape} presented an efficient data-driven EM algorithm to iteratively optimize shape alignment and matching correspondences. Felzenszwalb et al.~\cite{ShapeTree} proposed to hierarchically match shapes using the dynamic programming algorithm, demonstrating good potential in capturing large shape deformations. An MCMC-based sampling algorithm was discussed in~\cite{LinGraphMatch} to solve multi-layer shape matching.
 To overcome the problems caused by incomplete or noisy contours, Zhu et al.~\cite{ShiShapeECCV2008} presented a many-to-many contour matching algorithm using a voting scheme. Riemenschneider et al.~\cite{PartialMatchingECCV2010} solved the partial shape matching by identifying matches from fragments of arbitrary length to the reference contours.

%Despite the acknowledged success, these shape matching methods sometimes recognizing object categories have large intraclass variance.

%rely on prepared prototype templates or hand-tuned detection parameters.

%The main challenges for shape matching are: occlusion (i.e. missing of true contours of objects), incomplete contours (i.e. edges broken into pieces), and confusion to backgrounds (i.e. true contours of objects connected to background clutters). Therefore, many methods address these problems by designing robust shape descriptors or matching scheme. For examples,

An alternative to shape detection is addressed by learning shape models for a given category of shape instances. These methods represent shapes as a loose collection of local contour fragments or an ensemble of pairwise constraints~\cite{ShottonPAMI08,LinICCV07,ShiShapeCVPR2010}. They usually involve the construction of a codebook of contour fragments (e.g. Groups of Adjacent Contours (GAS)~\cite{PAS}) and train the shape models by supervised leaning. For example, the boosting methods were employed to train the discriminative classifiers with contour-based features~\cite{ShottonPAMI08,ZissermanECCV2006}. Maji et al.~\cite{MalikCVPR2009} incorporated the Hough transform into a discriminative learning framework, in which the contour words and their spatial layout were optimized jointly. Kokkinos and Yuille~\cite{ShapeParsing} suggested hierarchically parsing shapes with the bottom-up and top-down computations, and adopted the multiple instance learning algorithm for model training. Another type of shape template is the active basis model proposed by Wu et al.~\cite{ActiveBasis}, which was trained with a shared sketch algorithm.

Very recently, major progress has been made in appearance-based object recognition using the latent structure models~\cite{LeoCCCP,LatentSVM,ShapeNIPS2012}, in which the latent variables effectively enrich the representations. These methods owe their success to their ability to cope with deformations, occlusions, and variations. Based on these methods, Srinivasan et al.~\cite{ShiShapeCVPR2010} trained the descriptive contour-based detector by using the latent-SVM algorithm, Song et al.~\cite{ContextSVMCVPR2010} integrated the context information with the SVM-based learning, and Schnitzspan et al.~\cite{HierachicalCVPR2009} further combined the latent discriminative learning with conditional random fields using multi-types of shape features.

The And-Or graph was originally explored by Zhu and Mumford~\cite{AOGgrammar} for modeling complex visual patterns. Its key idea, using And/Or nodes to account for structure reconfigurations and variabilities in hierarchical composition, has been extensively applied in several vision tasks such as object and scene parsing~\cite{LeoAOG,HierarchicalPoslets,LinGrammar} and event analysis ~\cite{EventGrammar}. However, these approaches often require elaborate annotations or manual initializations. Si and Zhu~\cite{AOTemplate}  recently presented a framework for unsupervised learning of the And-Or image template, and demonstrated very promising results on modeling complex object categories. Our approach is partially motivated by these works, and we target on an alternative way to discriminatively train the And-Or graph model with the non-convex optimization. Our preliminary attempts along this path have been discussed in \cite{ShapeNIPS2012,LinAndOrTree}.

%Our approach demonstrates superior performance on very challenging cases for shape detection and localization. {\bf new dataset...}


%\vfill


\section{Representations}\label{sec:representation}

In this section, we define all the components of our And-Or graph models, including the shape features and the potential functions for graph nodes and edges.

\subsection{Contour Descriptor}

%For each decomposed block part, only one contour is selected as the proposed detection result.

\begin{figure}[!htb]
\centering
\epsfig{figure=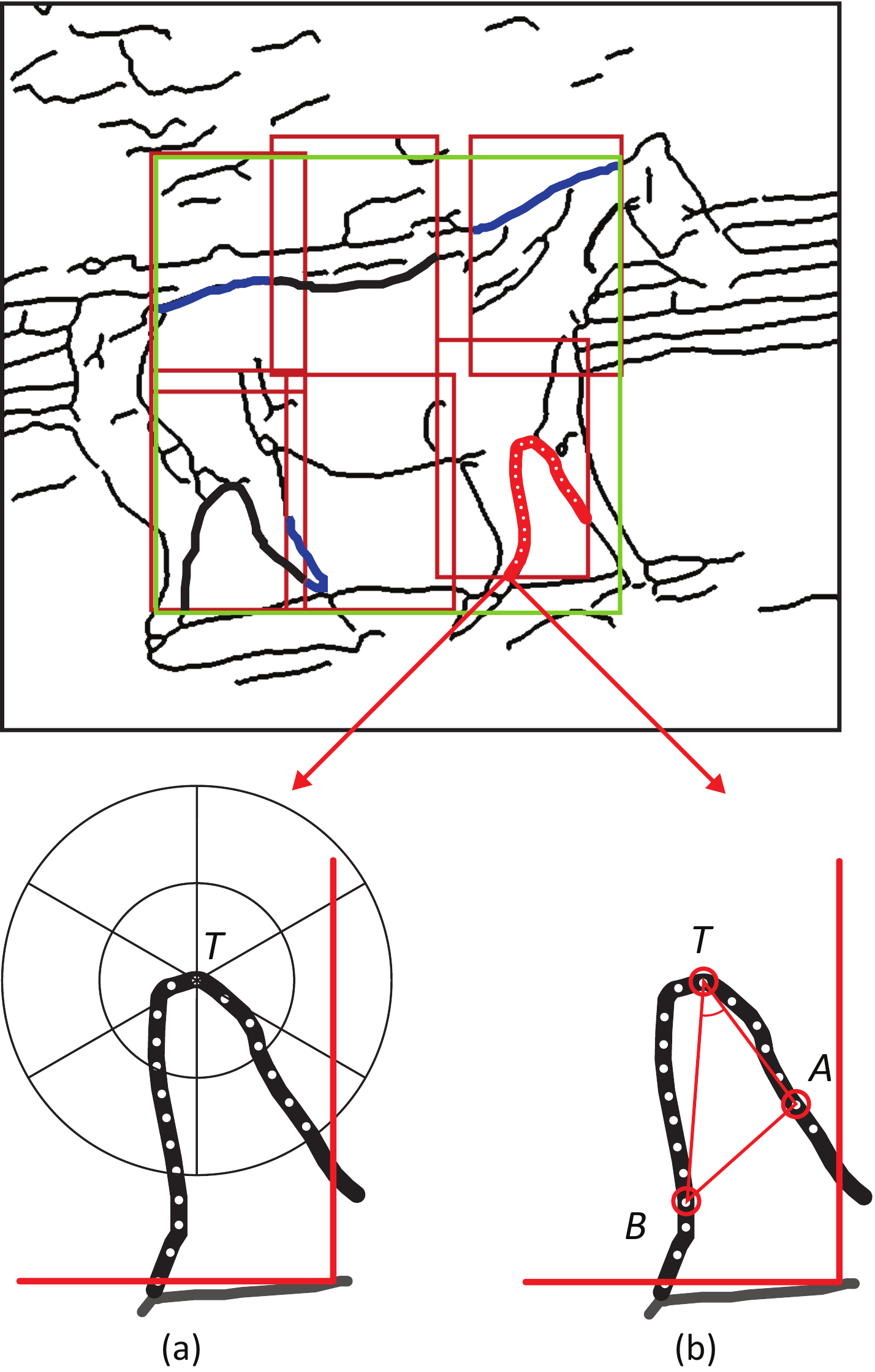, width=2.1 in}
\caption{ Illustration of the proposed contour descriptor. This feature combines the Shape Context descriptor in (a)  and the triangle-based descriptor in (b) to characterize a local contour fragment. }\label{fig:feature}
\end{figure}

First, we introduce our contour descriptor for characterizing local contour fragments. As Figure~\ref{fig:feature} illustrates, this feature combines the triangle-based descriptor~\cite{ShapeGroup} and the Shape Context~\cite{ShapeContext}, capturing local contour deformations with the surrounding contexts. For any contour fragment we extract a sequence of sample points $\Omega$, and for each point in $\Omega$, its triangle-based descriptor and Shape Context descriptor are both computed and concatenated into a vector. Then we pool the vectors of all the sample points into a histogram.

Given a point $T \in\Omega$ for a contour, we collect triangles that are formed by $T$ and any other two $A, B$ in $\Omega$. Note that each triangle is constructed by three different points. As Figure~\ref{fig:feature}(b) illustrates, the triangle-based descriptor for $T$ is a 3-D histogram, denoted by $\mathbb{H}^{t}(T)$, which contains the angle values (e.g. $\angle{BTA}$) and the two distances $TA$ and $TB$ in each dimension,  respectively. We use the clockwise orientation to determine the triangle $\angle{BTA}$, and the distances $TB$ and $TA$ are normalized by the average distance between the points in $\Omega$. The Shape Context descriptor, denoted by $\mathbb{H}^{b}(T)$, is constructed by $T$ and all other points in $\Omega$.

%Given a point $T \in\Omega$ of a contour, the triangle-based descriptor is a histogram based on the triangles constructed by $T$ and any other two points $A, B \in \Omega$ (Figure~\ref{fig:feature} (b)). More precisely, it is a 3D histogram to represent the angles $\angle{BTA}$, and the two distances $TA$ and $TB$, respectively. Note that triangle $\angle{BTA}$ is calculated by clockwise orientation and distances $TB$ and $TA$ are normalized by the average distance between points in $\Omega$.  For the Shape Context descriptor, it counts for the lengths and polar angles of the vectors from $T$ to all other points in $\Omega$.

In our implementation, the number of sample points for each contour fragment is fixed at $20$ , and the distances between adjacent points in $\Omega$ are equal. For each point $T$, $(20-1)*(20-2) / 2 = 171$ triangles are thus collected.  We define the 3-D histogram $\mathbb{H}^{t}(T)$ including 2 bins for $TA$, 2 bins for $TB$, and 6 bins for angle $\angle{BTA}$ ranging from $0$ to $\pi$. We transform $\mathbb{H}^{t}(T)$ into a $2 \times 2 \times 6 = 24$-bin 1-D feature vector. For the Shape Context descriptor $\mathbb{H}^{b}(T)$, we use $2$ bins for lengths and $6$ bins for polar angles ranging from $0$ to $2\pi$, then its length is $2 \times 6 = 12$. By concatenating these two descriptors, we obtain the feature vector of $T$ including $(24+12) = 36$ bins. Thus, the contour fragment is represented by a feature vector of $36 * 20 = 702$ bins.

\subsection{And-Or Graph Model}

Our model is defined in the form of an And-Or graph $\mathcal{G}= (\mathcal{V},\mathcal{E})$, where $\mathcal{V}$ includes four levels of nodes and $\mathcal{E}$ includes the graph edges. The root-node is indexed as 0, indicating the switch among different shape views,(or other different global variations, by analogy). The and-nodes are indexed by $r = 1,...,m$, with each representing one global classifier. For each and-node, there are a number of $z$ or-nodes arranged in a layout of $b_1 \times b_2$ blocks to represent several object parts, and we index all of the or-nodes as $j = m + 1, ..., (z+1) * m$. The leaf-nodes in the fourth layer are indexed by $i = (z+1)*m+1,...,(z+1)*m+1+n$, where $n$ is the number of leaf-nodes. For notation simplicity, we define $m^{\prime} = (z+1)*m+1, n^{\prime} = (z+1)*m+1+n$, and $i \in ch(j)$ indicating a child node of node $j$. The details of the model $\mathcal{G}$ are described as follows.

%The proposed shape model is constructed in the form of an And-Or graph representation $\mathcal{G}= (\mathcal{V},\mathcal{E})$, where $\mathcal{V}$ represents three types of nodes and $\mathcal{E}$ the graph edges. As Figure~\ref{fig:AOG_example} illustrates, the square on the top is the root-node capturing the global discrimination of the shape category. The dashed circles derived from the root are a number of $z$ or-nodes arranged in a layout of $a_1 \times a_2$ blocks, representing the parts of object shapes. Each or-node comprises an unfixed number of leaf-nodes, which are allowed to be automatically adjusted during model training. For notation simplicity, we set the maximum number of leaf-nodes is $m$ corresponding to each or-node. Then the maximum number of all nodes in the model is $ 1 + n = 1 + z + m \times z $. We use $i = 0$ indexes the root node, $i = 1,...,z$ the or-nodes and $j = z + 1,...,n$ the leaf-nodes. We also define that $j \in ch(i)$ denotes child node $j$ associating to parent node $i$. The horizontal graph edges (i.e. collaborative edges) are defined between the leaf-nodes that are derived from different or-nodes. That is, no edges are connected among leaf-nodes belong to the same parent or-node. In this way, the collaborative edges encode the compatibilities among shape parts. The details of the model $\mathcal{G}$ are presented as follows.

\textbf{Leaf-node:} Each leaf-node $L_i $ is a local classifier for detecting partial shape contours. We denote the location of leaf-node $L_i$ as $p_i$, which is determined by its parent or-node. Given the extracted edge map $X$, we treat contour fragments within the observed block as the inputs of $L_i$. For a contour $c$, we denote $\phi^{l}(p_i,c)$ as its feature vector using the proposed contour descriptor, and only the part of $c$ that has fallen into the block will be considered. Note that we can prune some very short contours as noises in practice. The response of classifier $L_i$ located at $p_i$ is defined as:

\begin{equation}\label{eq:leaf-node_score}
 \mathcal{R}_{i}^{l}(X,p_i) =\max_{c \in X} \omega_{i}^{l} \cdot \phi^{l}(p_i,c),
\end{equation}
where $\omega_{i}^{l}$ is a parameter vector that is set to zero if the corresponding leaf-node $L_j$ is nonexistent. We can thus localize the contour representing the shape part by $c_i = argmax_{c \in X} \omega_{i}^{l} \cdot \phi^{l}(p_i,c)$. This partial detecting scheme enables to partition true object contours from cluttered background.

\textbf{Or-node:} The or-node $U_j, j = m + 1, ..., (z+1) * m $ specifies one of its child leaf-nodes, and also the contour detected by the leaf-node. Every or-node is allowed to slightly perturb their locations with respect to the root in order to capture the inter-part deformations.

For each or-node $U_j$, we define the deformation feature, $\phi^{s}(p_0, p_j) = (dx,dy,dx^2,dy^2)$, where $(dx,dy)$ encodes the displacement of the or-node position $p_j$ to the expected position $p_0$ determined by the root-node. The cost of locating $U_j$ at $p_j$ is:
\begin{equation}
 D_{j}(p_0, p_j) = \omega_{j}^{s} \cdot \phi^{s}(p_0, p_j),
\end{equation}
where $\omega_{j}^{s}$ is a 4-dimensional parameter vector corresponding to $\phi^{s}(p_0, p_j)$.

For each leaf-node $L_i$ associated with $U_j$, we introduce an indicator variable $v_i \in \{ 0, 1 \}$ representing whether it is activated by $U_j$ or not. We also define an auxiliary vector for $U_j$, $\mathbf{v}_j = \{v_{i}\}_{i \in ch(j)}$, where $ ||\mathbf{v}_{j}|| = 1$ or $0$. Note that $||\mathbf{v}_{j}||=1$ only when one of the leaf-nodes  under $U_j$ is activated. In this way, the or-node can adaptively activate the different leaf-nodes to capture the diverse local shape variance. It is worth mentioning that  the cost of locating the or-node is independent of the selected leaf-nodes because we assume the leaf-nodes belong to the same part (i.e. or-node) act a nearby location.

Thus, the response of the or-node $U_j$ is defined as,
\begin{equation} \label{eq:or-node_score}
\mathcal{R}_{j}^{u}(X,p_0,p_j,\mathbf{v}_{j}) \!\! = \!\!\! \sum_{i \in ch(j)} \mathcal{R}_{i}^{l}(X,p_j) \cdot v_i \!-\! D_{j}(p_0,p_j).
\end{equation}

\begin{figure}[!htb]
\centering
\epsfig{figure=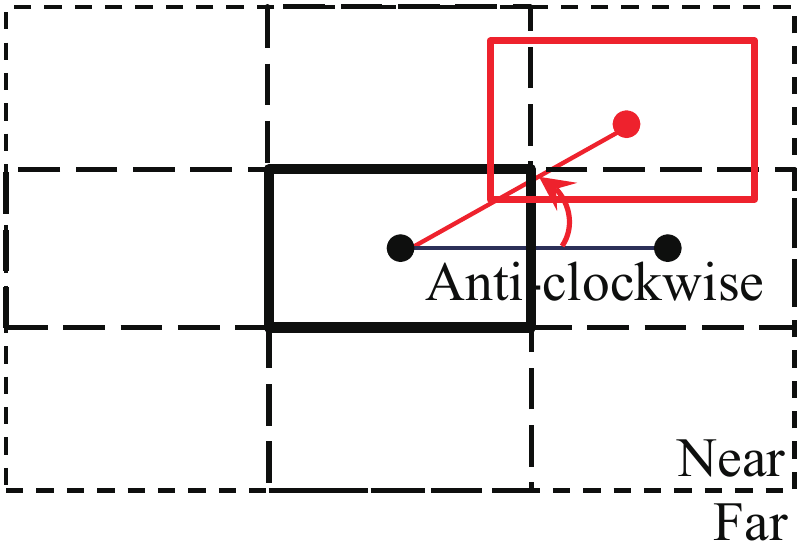,width=0.22\textwidth}
\caption{The spatial contextual features defined for the collaborative edges. }\label{fig:context_feature}
\end{figure}

\begin{figure*}[!htb]
\centering
\epsfig{figure=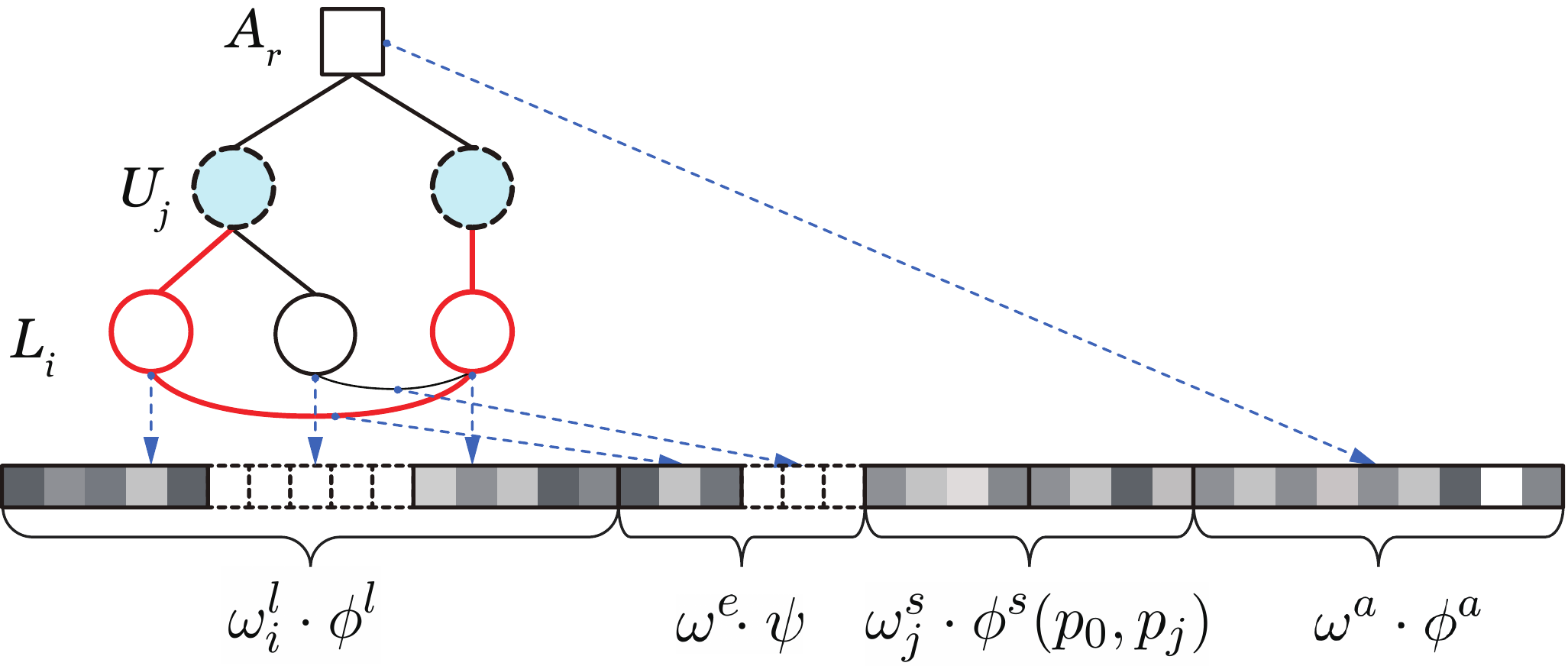,width=0.70\textwidth}
\caption{Mapping the latent And-Or graph with the discriminative function defined in Equation~\ref{eq:AOG_score}. Different layers of nodes in our model are associated with certain bins in the feature vector $\phi(X,H)$ (at the bottom). The activated leaf-nodes are highlighted in red, and the feature bins are set to zeros for the other inactivated nodes. The embedded latent variables  $H = (P, V)$  make our model reconfigurable during detection. }\label{fig:AoG_feature}
\end{figure*}

\begin{table*}[htbp]	
		\centering\begin{tabular*}{0.85\textwidth}{ p{0.2\textwidth} p{0.7\textwidth} l  l}
        \toprule
        Symbol                                        & Meaning \\
        \midrule
        $\{A_r\}_{r=1}^m$ & The and-nodes. \\
        $\{U_j\}_{j=m+1}^{ (z+1)\cdot m}$ & The or-nodes. \\
        $\{L_i\}_{i=m'}^{n'}$ & The leaf-nodes. \\
        $X$ & The edge map of an image. \\
        $P = \{p_0, p_j, p_i\}$ & The locations of the root-node $p_0$, or-nodes $p_j$, and leaf-nodes $p_i$. \\
        $\mathcal{R}^l_i(X, p_i)$ & The response of the classifier associated with leaf-node $L_i$ located at $p_i$. \\
        $\mathcal{R}^u_j (X, p_0, p_j, \mathbf{v}_j)$ & The response of the or-node. $\mathbf{v}_j$ indicates the selection of its child leaf-nodes. \\
        $\mathcal{R}^a_r(C_r)$ & The response of the and-nodes, which provides a global verification for the shape $C_r$. \\
        $\mathcal{R}^{e}_r (P, \{\mathbf{v}_j\}) $ & The response of the collaborative edges. $ \{\mathbf{v}_j\} $ indicates the selection of the leaf-nodes.\\
        $\mathcal{R}^G (X, P, V)$ & The response of the whole model, where $P$ and $V$ represent the latent variables. \\
        $H = (P, V)$ & All latent variables (including positions $P$ and activation variables $V$) of our model. \\
        \bottomrule
        \end{tabular*}
		\vspace{3mm}
		\caption{Notation summary of this work.}\label{table:notation}
\end{table*}

\textbf{And-node:} The and-node, $A_r$, performs a global verification for the whole shape. For each and-node, we have a set of contour fragments, $C_r=\{c_1, c_2, ...,c_z\}$, which are determined by its child or-nodes. Then we  adopt the Shape Context descriptor~\cite{ShapeContext} to describe these contours as a whole, $\phi^{a}(C_r)$. Thus, we define the and-node's response as,
\begin{equation} \label{eq:and_score}
 \mathcal{R}^{a}_{r} (C_r) = \omega^{a} \cdot \phi^{a}(C_r),
\end{equation}
where $\omega^{a}$ is the corresponding parameter vector.

\textbf{Collaborative Edge:} We impose contextual interactions among shape parts based on the collaborative edges. Given any two different or-nodes associated with the same and-node, we link an edge between them and their child leaf-nodes inherit the edge. 
We define the collaborative edges using the spatial contextual features, as Figure~\ref{fig:context_feature} illustrates.

Suppose one edge connects two leaf-nodes $(L_i, L_{i^{\prime}})$ are located at $p_i$ and $p_{i^{\prime}}$ respectively. We collect a $4$-bin feature $\psi (p_i, p_{i^{\prime}})$ for the two leaf-nodes according to their spatial layout. Each bin of $\psi (p_i, p_{i^{\prime}})$ represents one of the four relations of $(L_i, L_{i^{\prime}})$: {\em clockwise}, {\em anti-clockwise}, {\em near}, and {\em far}. In Figure~\ref{fig:context_feature}, the bold rectangle in the center indicates the location of $L_i$, which is connected to the red bold rectangle indicating the location of $L_{i^{\prime}}$. The dashed line represents the initial layout of the two leaf-nodes, and the red solid line is the adjusted actual layout in detection. Specifically, we define the relations as
\begin{itemize}
    \item {\em Near and Far}: If $L_{i^{\prime}}$ falls into the outer dashed rectangle, it is {\em near} to $L_i$, i.e. the bin of {\em near} is activated (i.e. being set as $1$); otherwise it is {\em far} from $L_i$.
    \item {\em Clockwise and Anti-clockwise}: One of the two relations is activated (i.e. being set as $1$) according to the angle between the dashed line and the solid red line.
\end{itemize}

These relations intuitively encode the spatial contexts of two leaf-nodes $(L_i, L_{i^{\prime}})$. Let $\{\mathbf{v}_j\}$ represent the activation variables of the leaf-nodes, and we denote $P$ as a vector of the positions of all or-nodes $U_j$. $P$ also specifies the locations $\{ p_i \}$ of the activated leaf-nodes.  The response of the collaborative edge is then parametrized as
\begin{equation}\label{eq:CRF_score}
 \mathcal{R}^{e}_r (P, \{\mathbf{v}_j\}) = \sum_{j \in ch(r)} \sum_{i \in ch(j)} \sum_{i^{\prime} \in \partial(i)} \omega_{(i,i^{\prime})}^{e} \cdot \psi (p_i, p_{i^{\prime}}) \cdot v_i \cdot v_{i^{\prime}},
\end{equation}
where $\partial(i)$ represents the set of neighbor leaf-nodes of $L_i$, and each neighbor has a different parent node with $L_i$. $\omega_{(i,i^{\prime})}^{e}$ is the corresponding weight. $v_i$ and $v_{i^{\prime}}$ are the activation indicators for $L_i$ and $L_{i^{\prime}}$, respectively, as the edges are imposed only for the activated leaf-nodes.

\begin{figure*}[!htb]
\centering
\epsfig{figure=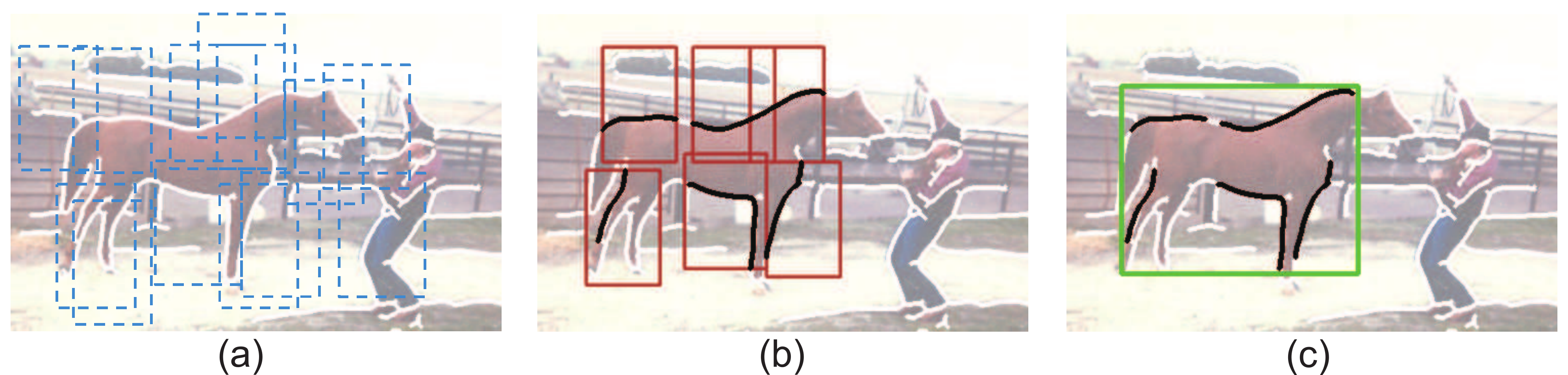,width=6.5in}
\caption{ Illustration of the inference procedure. (a) shows local testing for detecting contour fragments within the edge map; the blue dashed boxes represent perturbed blocks associated with the leaf-nodes. (b) shows a hypothesis of detection including candidates (indicated by the red boxes) proposed by all or-nodes, in which the collaborative edges are imposed. (c) shows the global verification, in which the ensemble of contours are measured as a whole.}\label{fig:inference}
\end{figure*}

\textbf{Root-node:} The root-node on the top alternatively activates one of its child and-nodes, whose definition is similar with that of the or-node. Also, we use a variable $v_r \in \{0,1\}$ to specify the activation of each and-node $A_r$, and the indicator vector for the root-node is $\mathbf{v}_0 = \{v_{r}\}_{r=1}^m$  and $||\mathbf{v}_0|| = 1$, i.e. only one child is selected.

Let $P$ imply the part-based deformation with or-nodes, and $V=(\mathbf{v}_0, \{\mathbf{v}_j\})$ imply the selection of and-nodes and leaf-nodes, the overall response of our model is then defined as:
\begin{eqnarray} \label{eq:AOG_score}
&&\mathcal{R}_{G}(X,P,V) = \nonumber\\  \!\! && \!\! \sum_{r=1}^{m} v_r \! \cdot \! ( \!\! \sum_{j \in ch(r)} \! \mathcal{R}^u_j(X, p_0, p_j,\textbf{v}_{j})  \!+ \! \mathcal{R}^e_r (P, \{\mathbf{v}_j\}) \!+\! \mathcal{R}^a_r(C_r) ).  \nonumber\\
\end{eqnarray}

In this model, $H = (P, V)$ are the latent variables that will be adaptively estimated in testing.  For notation simplicity, our model in Equation (\ref{eq:AOG_score}) can be re-written as :

\begin{equation}\label{eq:sim_score_infer}
 \mathcal{R}_{G}(X, H) = \omega \cdot \phi(X,H),
\end{equation}
where $\phi(X,H)$ represents the concatenated feature vector for all nodes and edges in the model, and $\omega$ includes all of the parameters corresponding to $\phi(X,H)$. Figure~\ref{fig:AoG_feature} illustrates our And-Or graph model mapped with the discriminative function.

We summarize the symbols used in our model in Table \ref{table:notation}.

\section{Inference}\label{sec:inference}

Given the edge map $X$ extracted from the image, the inference task is to detect the optimal contour fragments within the detection window scanned over an image pyramid. The detection is a search procedure to activate nodes from bottom to top, in which a number of hypotheses are generated and each one specifies a configuration of detected contour fragments. We verify the hypotheses and prune the unlikely ones by maximizing the model response defined in Equation (\ref{eq:sim_score_infer}).

%Assume that the root-node is located at $p_0$, and the hypothesis of detection is denoted as  $\hat{H} $, including all or-nodes' locations and their selections to activate the child leaf-nodes, and the root-node's selection on the top. The hypothesis will be verified in a bottom-up manner by maximizing the model response $ \mathcal{R}_{G}(X, H)$ defined in Equation (\ref{eq:sim_score_infer}).

%Assuming the root-node is located at $p_0$, we can detect the target shape by . We consider an valid hypothesis of detection as a composite set of these candidates,
%
%and it can be measured by the potential of collaborative edge in Equation (\ref{eq:CRF_score}). We denote each hypothesis as $\hat{H} = ( \{ \hat{p_j} \}, \{ \mathbf{\hat{v}}_{j} \})$, including the locations and activations of all or-nodes.

% we notate the vector of placements for root-node and or-nodes as $P=(p_0,p_{m+1},...,p_{(z+1)*m})$ and the vector of selections for and-nodes and leaf-nodes as $V=(v_1,...,v_m,v_{m^{\prime}},...,v_{n^{\prime}})$.

We conduct the inference algorithm with the following steps.  An example illustrating the inference procedure using our model is presented in Figure~\ref{fig:inference}.

\textbf{Local testing:} We use all of the leaf-nodes (i.e. the local contour classifiers) to search for optimal contour fragments within the edge map $X$. Assume that one or-node $U_j$, associated with a partitioned block in the detection window, contains a number of leaf-nodes $\{ L_i, i \in ch(j) \}$, and that the initial position of $U_j$ is $p'_j$. Each $U_j$ is allowed to slightly perturb its location. At each location $p'_j+ \delta $, we treat all of the contours that have fallen into the block as the inputs to every leaf-node of $U_j$, as Figure~\ref{fig:inference}(a) illustrates. By maximizing the response in Equation (\ref{eq:leaf-node_score}), each leaf-node $L_i \in ch(j)$ can find an optimal contour at a certain location. Recall that each or-node can activate only one of the child leaf-nodes. Thus, the possibility of different leaf-node selections can generate a batch of detection hypotheses. In particular, we denote $\hat{H}$ as the latent variables for one hypothesis, and denote $(\mathbf{\hat{v}}_{j}, \hat{p_j})$ for a possible activation of $U_j$, where $\mathbf{\hat{v}}_{j}$ indicates the leaf-node selection and $\hat{p_j}$ is the location. The cost of $(\mathbf{\hat{v}}_{j}, \hat{p_j})$ is then measured by the function $\mathcal{R}_{j}^{u}$ defined in Equation (\ref{eq:or-node_score}).

% and we notate $\mathbf{\hat{v}}_{j}$ as a hypothesis of activating one certain leaf-node. And we can obtain the optimal location of $U_j$ with respect to different hypotheses, by maximizing
%\begin{equation}\label{eq:placement}
%{p}^{*}_{j} = \arg\max_{p_j} \mathcal{R}_{j}^{u}(X,p_0,p_j,\mathbf{\hat{v}}_{j}).
%\end{equation}
%Accordingly, for each detection, the overall score of its local testing for a specific hypothesis $\mathbf{\hat{v}}_{j}$ is calculated by assembling all the responses of its parts, i.e. the or-nodes, defined in Equation (\ref{eq:or-node_score}).

%\begin{equation}\label{eq:BotUp}
%S_r^l(X, P^*, \mathbf{\hat{v}}_j) = \sum_{j \in ch(r)} \mathcal{R}_{j}^{u}(X,p_0,{p}^{*}_{j}, \mathbf{\hat{v}}_{j}).
%\end{equation}

\textbf{Binding testing:} The hypotheses from the local testing are further weighed and filtered by imposing the collaborative edges. In each hypothesis, each or-node proposes one leaf-node, and any two leaf-nodes derived from different or-nodes are connected by an edge. We measure the score by the potential function in Equation (\ref{eq:CRF_score}).

%We denote each hypothesis as $\hat{H} = ( \{ \hat{p_j} \}, \{ \mathbf{\hat{v}}_{j} \})$, including the locations and activations of all or-nodes.

In this way, each detection hypothesis is scored by the two testing steps, as,
\begin{eqnarray}\label{eq:testing}
S_{r}^{l}(X, \hat{H}) = \sum_{j \in ch(r)} \mathcal{R}_{j}^{u}(X,p_0,\hat{p_j}, \mathbf{\hat{v}}_{j}) +   \mathcal{R}^{e}_r (\hat{P}, \{ \mathbf{\hat{v}}_j \}),
\end{eqnarray}
where $\hat{P} = \{ \hat{p_j} \}$ denotes the locations of all of the or-nodes. In practice, we can prune some of the hypotheses by setting a threshold on the score.

\textbf{Global verification:}
In this step, we apply the and-nodes to re-score the hypotheses of detection. For any hypothesis, we obtain an ensemble of contours, $\hat{C_r} = \{ \hat{c}_1, \hat{c}_2, \dots, \hat{c}_z \}$, each of which is proposed by one or-node. We can measure the contours as a whole by $ S_{r}^{g} (X,\hat{H}) = \mathcal{R}^{a}_{r} (\hat{C_r})$ in Equation (\ref{eq:and_score}), as Figure \ref{fig:inference}(c) illustrates.

Afterwards, the root-node determines the optimal detection by selecting the maximum aggregated score, as
\begin{equation}\label{eq:global}
H^* = \arg\max_{\hat{H}} \sum_{r=1}^{m} (S_{l}(X,\hat{H}) + S_{r}^{g} (X,\hat{H}) )\cdot \mathbf{\hat{v}}_0,
\end{equation}
where $|| \mathbf{\hat{v}}_0 || = 1$ constrains only one of the and-nodes selected by the root-node.

%For all hypotheses $V_j$ generated by the two steps of testing, we further verity them by integrating the previous two stages together. Let $\mathbf{\hat{v}}_0$ denote the hypothesis of and-node selection, and ${V}=\{\mathbf{\hat{v}}_0, \mathbf{\hat{v}}_j\}$ denote all the hypotheses of the selection of both and-nodes and leaf-nodes. The optimal hypothesis can be obtained by maximizing the overall response of the graph defined in Equation (\ref{eq:AOG_score}), which can be formulate as follows,

%And the detection result is obtained by select the optimal $H^* = (P, V)$ which gives the highest overall response by sliding window scheme.

The overall inference procedure appears in Algorithm \ref{alg:inference}.

%%%%%%%%%%%%%%%%%%%%%%%%%%%%%
%% Algorithm flowchart :  INFERENCE
%%%%%%%%%%%%%%%%%%%%%%%%%%%%%
\begin{small}
\begin{algorithm}[htb]
\caption{Inference with the And-Or graph representation}
\label{alg:inference}
\begin{algorithmic}\footnotesize
\REQUIRE ~~\\ % INPUT
    $X$: the edge map extracted from the test image.
\ENSURE ~~\\    % OUTPUT
    $H^*$: the optimal detection with the maximal detection score $\mathcal{R}_{G}(X, H^*)$.

\vspace{1.3em}
 \textbf{Local testing}: 
 \STATE 1. Apply leaf-nodes to detect all possible local contour fragments.
 \STATE 2. Generate a batch of detection hypotheses via the or-nodes.

%Measure each candidate $(\mathbf{\hat{v}}_{j}, \hat{p}_j)$ by the potential $\mathcal{R}_{j}^{u}$ defined in Equation  (\ref{eq:or-node_score}).
%%\begin{equation}
%%\mathcal{R}_{j}^{u}(X,p_0,\hat{p}_j,\mathbf{\hat{v}}_{j}) \!\! = \!\!\! \sum_{i \in ch(j)} \mathcal{R}_{i}^{l}(X,p_j) \cdot \hat{v}_i \!-\! D_{j}(p_0,\hat{p}_j). \nonumber
%%\end{equation}
\textbf{Binding testing}: 
\STATE 1. Impose the collaborative edges between leaf-nodes in each detection hypothesis.
\STATE 2. Score the hypotheses by Equation (\ref{eq:testing}).
\STATE 3. Prune unlikely hypotheses by thresholding the score. 

\textbf{Global verification}:
\STATE 1. For each hypothesis, the local contours are measured as a whole via the and-nodes. 
\STATE 2. Aggregate all potentials via the root-node in Equation (\ref{eq:global}). 
\STATE 3. Merge results by non-maximum suppression over all image positions and scales.

\end{algorithmic}
\end{algorithm}
\end{small}

\section{And-Or Graph Learning}\label{sec:learning}

\begin{figure*}[!htb]
\centering
\epsfig{figure=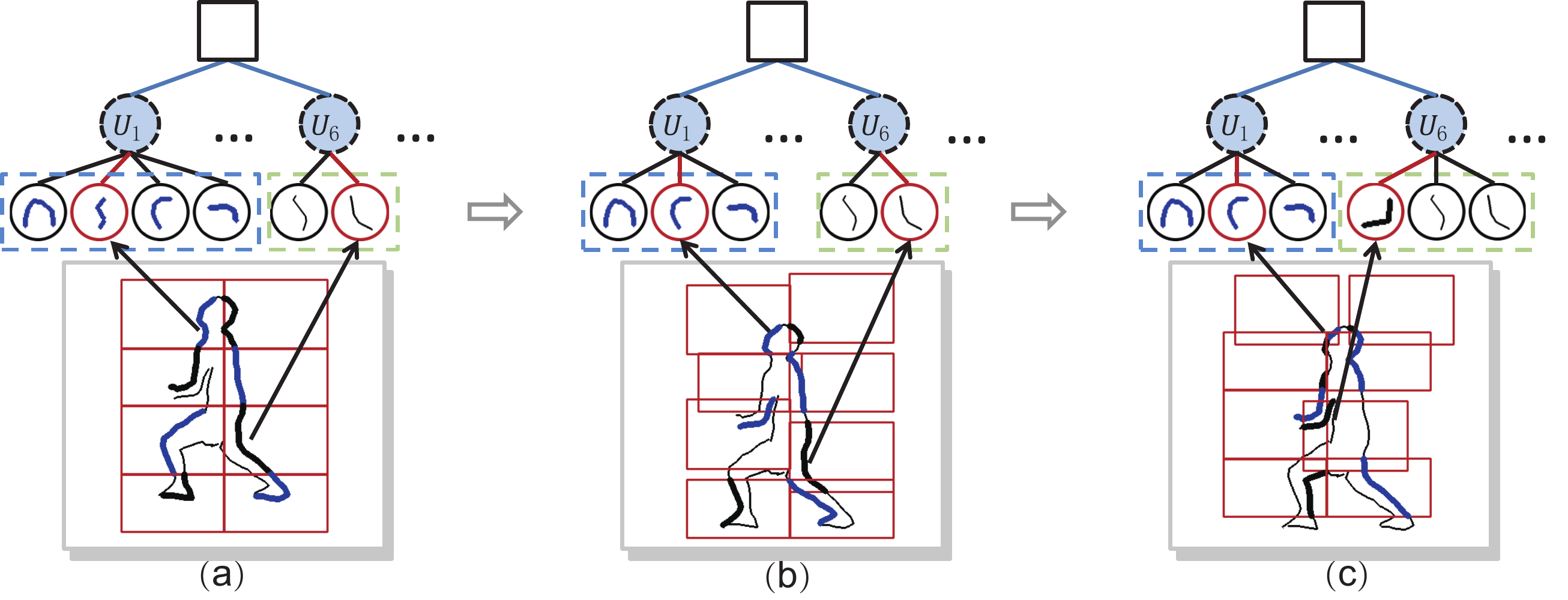,width=0.9\textwidth}
\caption{Illustration of the structure reconfiguration. Parts of the model, two or-nodes ($U_1,U_6$), are visualized in three intermediate steps. (a) The initial structure, i.e. the regular layout of an object. Two new structures are dynamically generated during the iterations. (b) A leaf-node associated with $U_1$ is removed. (c) A new leaf-node is created and assigned to $U_6$.
 }\label{fig:structure}
\end{figure*}

We formulate the And-Or graph model learning as a joint optimization task of model structures and parameters. To achieve this goal, we present a novel structure learning algorithm extended from the existing non-convex optimization methods~\cite{CCCP,ShapeNIPS2012}. This algorithm optimizes the objective in a dynamical manner: the latent structures $H = (P,V)$ are iteratively determined along with the parameter learning in each step. For example, new leaf-nodes are created or removed to better adapt to the training data by adjusting the latent variables. One instance of our learning procedure is illustrated in Figure~\ref{fig:structure}: from (a) to (b), a leaf-node associated with $U_1$ is removed and a new leaf-node under $U_6$ is created in (c).

\subsection{Optimization Formulation}

Suppose we have a set of positive and negative training samples $(X_1,y_1)$,...,$(X_N,y_N)$, where $X$ is the edge map and $y=\pm 1$ is the label indicating positive and negative samples. We assume that the samples indexed from $1$ to $K$ are the positive samples, and that the feature vector for each sample $(X,y)$ is,
\begin{equation}\label{eq:feature_phi}
 \qquad \phi(X,y,H) =
  \left\{
   \begin{array}{lr}
   \phi(X,H) & \mbox{if } y = +1 \\
   0 & \mbox{if } y = -1\\
   \end{array}
  \right.,
\end{equation}
where $H$ represents the latent variables and $\phi(X,H)$ the overall feature vector of the And-Or graph model. Then we pose the And-Or graph learning as optimizing model parameters along with the latent structures,
\begin{equation} \label{eq:discriminative_fun}
 \omega = argmax_{y,H} (\omega \cdot \phi(X,y,H)).
\end{equation}
We further transfer this target into a maximum margin formulation,
\begin{align} \label{eq:learn_opt}
\min_{\omega} \frac{1}{2} \| \omega \|^2  & +  \lambda \sum_{k=1}^N[\max_{y,H}(\omega \cdot \phi(X_k,y,H) + \mathcal{L}(y_k,y,H)) \nonumber\\
& -  \max_H (\omega \cdot \phi(X_k,y_k,H))],
\end{align}
where $\lambda$ is a penalty weight (set as 0.005 empirically), and $\mathcal{L}(y_k,y,H)$ is the loss function. In our implementation, we define that $\mathcal{L}(y_k,y,H) = 0$ if $y_k = y$, and 1 otherwise.

The target energy in Equation (\ref{eq:learn_opt}) is non-convex making it difficult to be solved analytically. In this work, we propose the Dynamical Structural Optimization (DSO) method to iteratively optimize this objective based on the Concave-Convex Procedure (CCCP) method~\cite{CCCP}.

\subsection{Dynamical Structural Optimization}

\begin{figure}[!htb]
\centering
\epsfig{figure=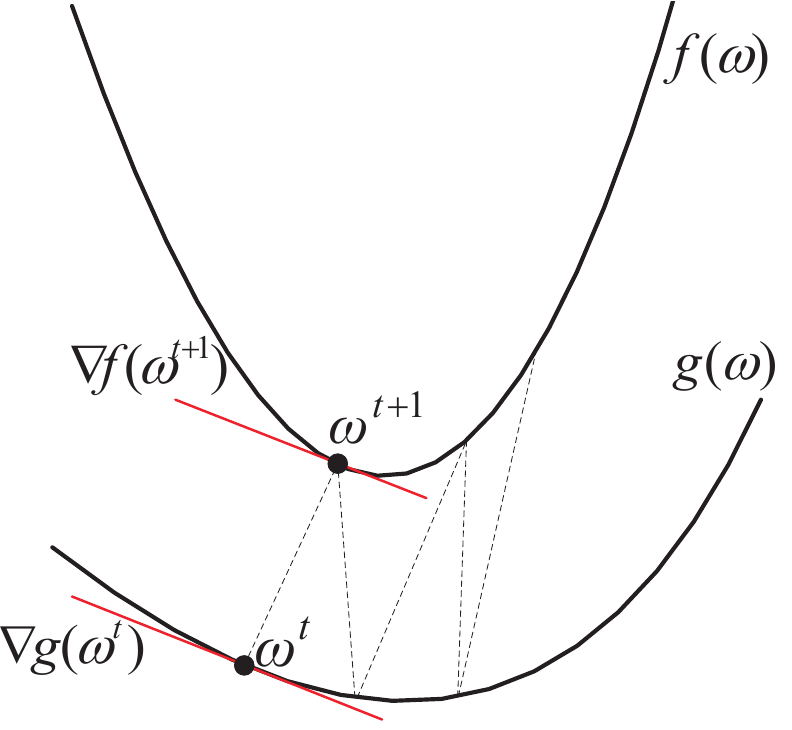,width=0.35\textwidth}
\caption{Geometric illustration of the CCCP procedure. The target energy is decomposed into two functions, $f(\omega)$ and $g(\omega)$. At each step of iteration, a hyperplane (represented by the red line) is calculated as the upper-bound at $\omega^t$ for optimizing $\omega^{t+1}$. }\label{fig:CCCP}
\end{figure}

Following the CCCP method~\cite{CCCP}, we convert the objective function in Equation  (\ref{eq:learn_opt}) into a convex and concave form as,
\begin{align}\label{eq:cccp_f}
\quad \min_{\omega}[ \frac{1}{2} \|\omega \|^2
& + \lambda \sum_{k=1}^N \max_{y,H}(\omega \cdot \phi(X_k,y,H) + \mathcal{L}(y_k,y,H))] \nonumber\\
& -  [\lambda \sum_{k=1}^N \max_{H} (\omega \cdot \phi(X_k,y_k,H))] \\
& =  \min_{\omega} [f(\omega) - g(\omega)],\label{eq:opt_target}
\end{align}
where $f(\omega)$ represents the first two terms, and $g(\omega)$ represents the last term in (\ref{eq:cccp_f}). Assume $\omega^t$ is the solution for the $t$-th iteration. The solution $\omega^{t+1}$ for the next iteration can be solved by subjecting it to
\begin{equation}
  \nabla f(\omega^{t+1}) = \nabla g(\omega^t).
\end{equation}
A geometric explanation of CCCP is presented in Figure~\ref{fig:CCCP}, where $\nabla g(\omega^t)$ can be regarded as a hyperplane (the red line) at $\omega^t$ (the black spot) to upper bound $- g(\omega)$. $\nabla g(\omega^t)$ can be solved analytically once $H$ is fixed. Then, the $\omega^{t+1}$ can be estimated accordingly by minimizing $f(\omega^{t+1})$. Please refer to~\cite{CCCP} for the theoretical background.

During the training procedure, the model parameters $\omega^t$ and latent structures $H^t$ are iteratively updated. To discover the models structures, we add one step called model reconfiguration in each iteration. Recall that the model structures (e.g. graph nodes) are mapped with the feature vectors, as Figure~\ref{fig:AoG_feature} illustrates. In this step, from the feature vectors of all positive training examples, we first extract the sub-vectors that are corresponding to different nodes (i.e. and-nodes or leaf-nodes), and each node, we perform clustering on these sub-vectors, respectively. Then, according to the clustering results, we rearrange each feature vector by placing the sub-vectors back into the feature vectors (e.g. re-assigning contour fragments to leaf-nodes). Consequently, the new model structures can be generated. Our DSO method iteratively performs with three following steps: (i) estimate the latent variables of training samples; (ii) reconfigure the model structures; (iii) update model parameters for the new structures.

{\bf (I)} The model parameters $\omega_t$ in the previous iteration are fixed. We find a hyperplane $q_t$ to upper bound $-g(\omega)$ in Equation (\ref{eq:opt_target}),
\begin{equation}\label{eq:CCCP_cons}
 -g(\omega) \leq -g(\omega_t) + (\omega-\omega_t) \cdot q_t, \forall \omega.
\end{equation}
The optimal latent variables $H_k^{*}$ are specified for each positive training example by,
\begin{equation}\label{eq:latentH_comp}
H_k^* = argmax_{H} (\omega_{t} \cdot \phi(X_k,y_k,H)).
\end{equation}
Note that we only take the positive training examples into account as $\phi(X_k,y_k,H) = 0$ when $ y_k = -1$. That is, we apply the current model to perform detections on the training samples, and the hyperplane is constructed as
\begin{equation}\label{eq:hyperplane}
q_t = - \lambda \sum_{k=1}^N \phi(X_k,y_k,H_k^*).
\end{equation}

%The original CCCP includes two iterative steps: (I) fixing the model parameters, estimate the latent variables $H$ for all positive samples; (II) compute the model parameters given $H$. In our method, besides the inferred $H^*$, we need to further determine the graph configuration, i.e. the production of leaf-nodes associated with or-nodes, to obtain the complete structure. Thus, we insert one step between two original ones to perform the structure reconfiguration. The three iterative steps are presented as follows.

%{\bf (I)} For optimization, we first find a hyperplane $q_t$ to upper bound the concave part $-g(\omega)$ in Equation (\ref{eq:opt_target}),
%\begin{equation}\label{eq:CCCP_cons}
% -g(\omega) \leq -g(\omega_t) + (\omega-\omega_t) \cdot q_t, \forall \omega,
%\end{equation}
%where $\omega_t$ includes the model parameters obtained in the previous iteration. We construct $q_t$ by calculating the optimal latent variables $H_k^* = argmax_{H} (\omega_{t} \cdot \phi(X_k,y_k,H))$. Since $\phi(X_k,y_k,H) = 0$ when $ y_k = -1$, we only take the positive training samples into account during computation. Then the hyperplane is constructed as $q_t = - \lambda \sum_{k=1}^N \phi(X_k,y_k,H_k^*)$.

{\bf (II)}
 In the second step, we optimize the model structures based on the estimated latent variables $H^*$. All graph nodes in our model are mapped with several feature bins (i.e. sub-vectors) of $\phi(X_k,y_k,H_k^*)$ for all of the training samples, as Figure~\ref{fig:AoG_feature} illustrates. Hence, we achieve the model reconfiguration process by rearranging $\phi(X_k,y_k,H_k^*)$. For example, we can remove leaf-node $L_j$ by setting the corresponding bins for $L_j$ into zeros. Specifically, two sub-steps are sequentially performed to generate and-nodes and leaf-nodes, respectively.

 {\bf (i) Global structure reconfiguration.} In the layer of and-nodes, we perform clustering on the feature vectors corresponding to the and-nodes, i.e. the global shape features defined in Equation (\ref{eq:and_score}). Note that each vector is a part of $\phi(X_k,y_k,H_k^*)$. The training object shapes detected by the same and-node are initially grouped into one cluster. We then perform clustering on all of the feature vectors by using ISODATA with Euclidean distance. Based on the clustering result, we rearrange the feature vectors mapping with the and-nodes. For example, if one vector is grouped into a new cluster $\Omega_r$, we shall move it into the bins corresponding to And-node $A_r$, and set its original bins as zeros.  In our implementation, we fix the number of and-nodes as $m$, to simplify the computation.

% Note that we can easily relax this restriction to unfixed number of views in more complex situations.

  {\bf (ii) Local structure reconfiguration.} After the global structure reconfiguration, each and-node is associated with a group of training examples. Suppose the and-node $A_r$ includes a number of or-nodes, and every or-node $U_j, j \in ch(r)$ further derives its child leaf-nodes $L_j, i \in ch(j)$. In this step, we configure the part-level structures rooted by $U_j$. Note that this step processes each or-node and its leaf-nodes separately.

Each or-node $U_j$ specifies one part of the whole object shape. Given the training examples associating with $A_r$, we extract the local contour features from $\phi(X_k, y_k, H_k^{\prime})$, which are corresponding to the shape part of $U_j$. Then we perform clustering on these vectors, and rearrange these vectors in $\phi(X_k, y_k, H_k^{\prime})$, similarly as the operation on the and-nodes. In our implementation, the number of leaf-nodes is not fixed, as the local variances of shapes are usually unpredictable. Thus, there are two specific operators to generate the leaf-nodes according to the clustering.

\begin{itemize}
    \item One new leaf-node is created if an extra cluster is generated.
    \item One leaf-node is removed if there are very few samples in the corresponding cluster.
\end{itemize}

\begin{figure}[!htb]
\centering
\epsfig{figure=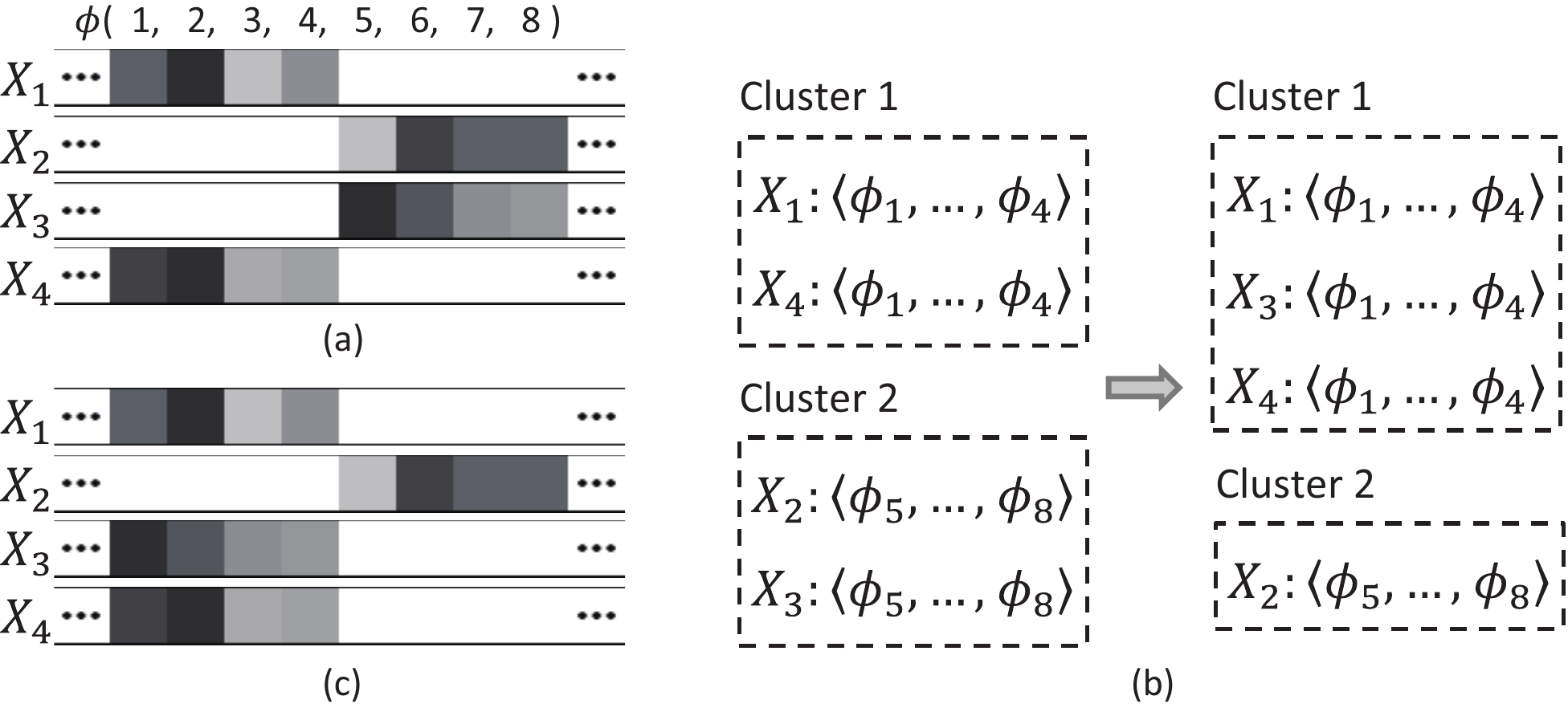,width=0.48\textwidth}
\caption{A toy example for structure reconfiguration. We consider $4$ samples, $X_1, \ldots, X_4$, for training the structure of $U_i$ (or $A_r$). (a) shows the feature vectors $\phi$ of the samples associated with $U_i$ (or $A_r$), and the intensity of the feature bin indicates the feature value. (b) illustrates the clustering performed with $\phi^{\prime}$. The vector $\langle \phi_5, \cdots, \phi_8, \rangle$ of $X_2$ is grouped from cluster 2 to cluster 1. (c) shows the adjusted feature vectors according to the clustering. Note that the model structure reconfiguration is realized by the rearrange of feature vectors, as we discuss in the text. This figure should be viewed in electronic form. }\label{fig:step2}
\end{figure}

We present a toy example in Figure~\ref{fig:step2} to illustrate the structure reconfiguration. For the sample $X_3$, a part of its feature vector $ < \phi_5, \ldots, \phi_8> $ is grouped from one cluster into another while the values of the feature bins are moved from $ < \phi_5, \ldots, \phi_8> $ to $ < \phi_1, \ldots, \phi_4> $.

%For each or-node $U_i$, we perform the clustering for detected contour fragments represented by the newly formed feature vectors. We first group the contours detected by the same leaf-node into the same cluster as a temporary partition. Then the re-clustering is performed by applying the ISODATA algorithm with Euclidean distance. And the close contours are grouped into the same cluster. According to the new partition, we can re-organize the feature vectors, i.e. represent the similar contour with the same bins in the complete feature vector $\phi$. Please recall that the vector of one contour is part of $\phi$.

 After the reconfiguration, the latent variables for each training example can be re-calculated, and denoted by $H_k^d$, in accordance with the arranged feature vectors (refer to Equation~(\ref{eq:latentH_comp})). We denote the feature vectors for all examples by $\phi^{d}(X_k,y_k,H_k^d)$. Then, the hyperplane is transformed accordingly, $q_t^{d} = - D\sum_{k=1}^N \phi^{d}(X_k,y_k,H_k^d)$.
 % Again, the main components of $q_t^{d}$ are preserved so that we consider $q_t^{d}$ approaches to $q_t$ and the convergence of optimization still exists.

%(c) shows the average precisions (AP) results generated by the And-Or tree (AOT) model and the And-Or graph (AOG) model.

{\bf (III)} The newly generated model structures can be represented by the feature vectors $\phi^{d}(X_k,y_k,H_k^d)$, and the model parameters can then be learned by solving Equation (\ref{eq:opt_target}),
\begin{equation}
\omega^d_{t} = argmin_\omega[f(\omega) - g(\omega)]
\end{equation}
By substituting $-g(\omega)$ with the upper bound hyperplane $q_t^{d}$, this optimization task can be transferred as,
\begin{align}
 \min_\omega \frac{1}{2} \|\omega\|^2 & +  D\sum_{k=1}^N[\max_{y,H}(\omega \cdot \phi(X_k,y,H) + \mathcal{L}(y_k,y,H)) \nonumber\\ 
& -  \omega \cdot \phi^d(X_k,y_k,H_k^{*})].
\end{align}
We solve it as a standard structural SVM problem, as,
\begin{equation}
 \omega^* = D \sum_{k,y,H} \alpha_{k,y,H}^* \Delta\phi(X_k,y,H),
\end{equation}
where $\Delta\phi(X_k,y,H) = \phi^d(X_k,y_k,H_k^{*}) - \phi(X_k,y,H)$. We calculate $\alpha^*$ by maximizing the dual form in standard SVM, and we apply the cutting plane method~\cite{CuttingPlane} to solve it. 

With the estimated parameters $\omega^d_{t}$, the energy $E(\omega^d_{t})$ can be calculated for the new model, and we then compare it with the previous energy $E(\omega_{t})$ to verify the new model structures. If $E(\omega^d_{t}) < E(\omega_{t})$, we accept the new model structures and have $\omega_{t+1} \leftarrow \omega^d_{t}$. Otherwise, we keep the model structures as in the previous iteration and optimize the model parameters without the structure reconfiguration, i.e. by using $q_t$ instead: $\omega_{t+1} = argmin_\omega[f(\omega) + \omega \cdot q_t]$.

In this way, we ensure that the optimization objective in Equation (\ref{eq:opt_target}) continues to decrease in iterations. Thus, the algorithm keeps iterating until the objective converges.

\subsection{Initialization}

At the beginning of model training, our model can be initialized as follows. For each training example, whose contours have been extracted, we partition it into a regular layout of partitioned blocks, and each block is corresponding to one or-node. The contours that fall into the block are treated as the inputs, and we initially select the one with the largest length if more than one contour are within there. Then, the leaf-nodes are initially generated by clustering the selected contours without any constraints. The and-nodes are initialized by the similar way. We thus obtain the initial feature vectors for all training examples.

Algorithm \ref{alg:Framwork} summarizes the overall algorithm of learning the latent And-Or graph.

%%%%%%%%%%%%%%%%%%%%%%%%%%%%%
%% Algorithm flowchart
%%%%%%%%%%%%%%%%%%%%%%%%%%%%%
\begin{small}
\begin{algorithm}[htb]
\caption{Learning latent And-Or graph model}
\label{alg:Framwork}
\begin{algorithmic}\footnotesize
\REQUIRE ~~\\
    positive and negative training samples, \\
    $\{X_k,y_k\}^{+} \bigcup \{X_{k^{\prime}},y_{k^{\prime}}\}^{-}, k = 1..K, k^{\prime} = K+1..N$.
\ENSURE ~~\\                           %�㷨��������Output
    The trained And-Or graph model.

\INPUT ~~\\
\begin{itemize}
\setlength{\itemsep}{1pt}
 \setlength{\parskip}{0pt}
 \setlength{\parsep}{10pt}

  \item[1] Initialize the model structure (the arrangement of nodes).
  \item[2] Initialize the latent variables $H$ and model parameters $\omega$.
\end{itemize}

\MYWHILE
    \STATE
    \begin{itemize}
\setlength{\itemsep}{1pt}
 \setlength{\parskip}{0pt}
 \setlength{\parsep}{10pt}
      \item[1] Estimate the latent variables $H^{*}$ on each positive example $(X_k,y_k)$ with the current model parameters $\omega_t$.
      \item[2] Generate the new graph structures.\\
        \begin{itemize}
\setlength{\itemsep}{1pt}
 \setlength{\parskip}{0pt}
 \setlength{\parsep}{10pt}
           \item[(a)] Localize the contour fragments for all examples using the current latent variables $H_k^{*}$, and obtain the feature vectors $\phi(X_k,y_k,H_k^{*})$.
           \item[(b)] In the layer of and-nodes, perform clustering on the global shape features, and rearrange the feature vectors.
           \item[(c)] For each or-node $U_i$, perform clustering on the feature vectors of all its child leaf-nodes.
           \item[(d)] Operate on the leaf-nodes to generate a new structure, and the latent variable is updated to $H_k^d$ with the rearranged feature vectors $\phi^d (X_k,y_k,H_k^d)$.
         \end{itemize}

      \item[3] Update the model parameters $\omega_{t+1}$ .
         \begin{itemize}
         \setlength{\itemsep}{1pt}
         \setlength{\parskip}{0pt}
         \setlength{\parsep}{10pt}
           \item[(a)] Estimate the parameters $\omega^d_{t}$ with the newly generated structures.
           \item[(b)] IF $E(\omega^d_{t}) < E(\omega_{t})$, 
           \item[]     ~~~~ Accept the new model structures, and $\omega_{t+1} \leftarrow \omega^d_{t}$.
           \item[] ELSE
           \item[]    ~~~~ Calculate $\omega_{t+1}$ while keeping the structures in the previous iteration.
          \end{itemize}

    \end{itemize}
\MYENDWHILE {The target function defined in Equation (\ref{eq:opt_target}) converges.}

\end{algorithmic}
\end{algorithm}
\end{small}

\begin{center}
\begin{table}[t]	
		\centering

\begin{tabular}{ c c c c c c }
  \hline
    & \textbf{SYSU-Shapes}  & \textbf{UIUC-People} &  \textbf{INRIA-Horses} & \\
     \textbf{and-nodes} & $m = 3 $ & $m = 2$   & $m = 1$ &    \\
     \textbf{or-nodes} & $z = 6$ & $z = 8$ & $z = 6$  &   \\
     \textbf{leaf-nodes}  & $n \leq 4$ & $n \leq 4$ & $n \leq 4$ &\\
     \hline
\end{tabular}
		\vspace{3mm}
		\caption{Numbers of nodes in the and-or graph models for different databases.}\label{table:graphnodes}
\end{table}
\end{center}

\section{Experiments}\label{sec:experiment}

To validate the advantage of our model, we present a new shape database, SYSU-Shapes\footnote{http://vision.sysu.edu.cn/projects/discriminative-aog/}, which includes elaborately annotated shape contours.  Compared with the existing shape databases, this database includes more realistic challenges in shape detection and localization, e.g. cluttered background, large intraclass variation, and different poses/views, in which part of the instances were originally used for appearance-based object detection. We also validate our model on two other public databases: UIUC-People~\cite{UIUCHuman} and INRIA-Horse~\cite{INRIAHorse} and show the superior performances over other state-of-the-art methods.

%We also evaluate our method on a newly proposed dataset constructed by us, namely XXX, to validate the effectiveness of our method.

\begin{figure*}[!htb]
\centering
\epsfig{figure=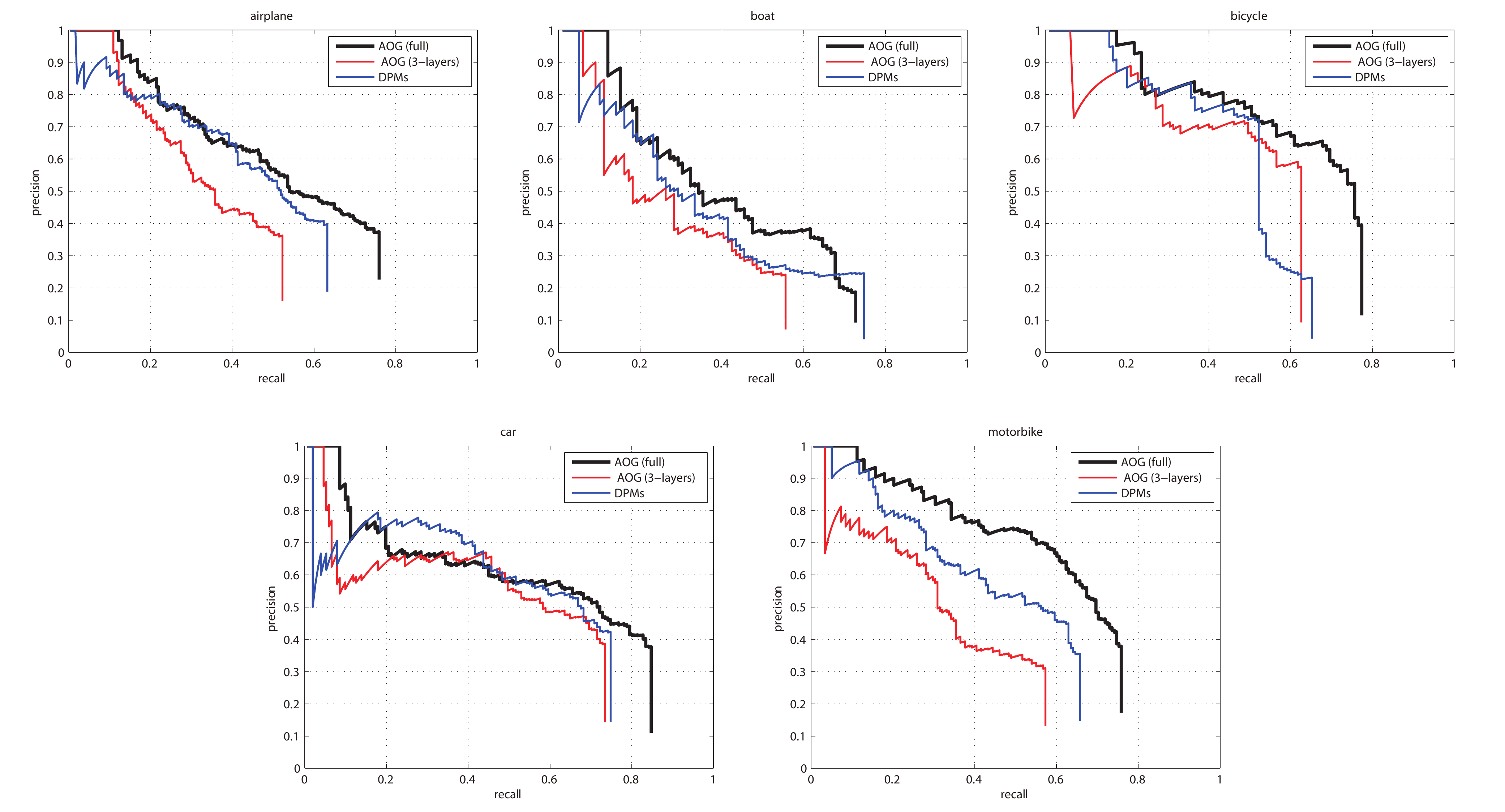,width=0.9\textwidth}
\caption{Precision-Recall (PR) curves on the SYSU-Shape dataset.}\label{fig:PR_SYSU_shape}
\end{figure*}

\begin{center}
\begin{table*}[t]	
		\centering\begin{tabular*}{0.55\textwidth}{ l c c c c c c} \toprule
         Method         & Airplane       & Bicycle        & Boat           & Car            & Motorbike      & MeanAP\\
        \hline
         AOG (full)     & \textbf{0.520} & \textbf{0.623 } & \textbf{0.419} & \textbf{0.549} & \textbf{0.583} & \textbf{0.539}\\
         AOG (3-layers)    & 0.348          & 0.482           & 0.288          & 0.466          & 0.333          & 0.383\\
         DPMs    & 0.437          & 0.488           & 0.365          & 0.509          & 0.455          & 0.451\\ \bottomrule
        \end{tabular*}
		\vspace{3mm}
		\caption{Detection accuracies on the SYSU-Shape dataset.}\label{table:SYSU_shape}
\end{table*}
\end{center}

{\em Implementation setting.}  We extract clutter-free object contours for the positive samples, and the edge maps for the negative samples are extracted using the Pb edge detector~\cite{PbDetector} with an edge link algorithm. For each contour as the input of the leaf-node, we sample $20$ points and compute the contour descriptor for each point. During detection, the edge maps of test images are extracted as for the negative training samples. The objects are searched by sliding windows over $6$ different scales and $2$ per octave, and detections are reported by non-maximum suppression.  We adopt the testing criterion defined in the PASCAL VOC challenge: a detection is counted as correct if its overlap with the groundtruth bounding-box is greater than $50\%$.

Our model is able to flexibly adapt to the data by setting the numbers of nodes in each layer: $m$ for and-nodes, $z$ for or-nodes, and $n$ for leaf-nodes. Recall that each or-node in our model indicates a part of object shape, so that we can set the number of or-nodes according to the sizes (scales) of the shape categories. The leaf-nodes are produced during the iterative training, and their numbers can be determined automatically. In the experiments, to reduce computational cost, we fix the number for and-nodes and set an upper limit for the number of leaf-nodes. Table~\ref{table:graphnodes} summarizes the numbers of nodes on the three databases. In the model training, the initial layout for each sample is a regular partition (e.g. $2 \times 4$ blocks for the UIUC-People dataset and $3 \times 2$ for the other two datasets).

If we keep only one and-node (i.e. $m=1$), our model is simplified into a 3-layer structure that is rooted by the and-node. The training procedure (i.e. Algorithm.~\ref{alg:Framwork}) for this structure is kept, but we discard the step of generating and-nodes.

%We notate this simplified version of our model as ``sim-AOG'', while the complete version as ``AOG'',  in the experiments.

\begin{figure*}[!htb]
\centering
\epsfig{figure=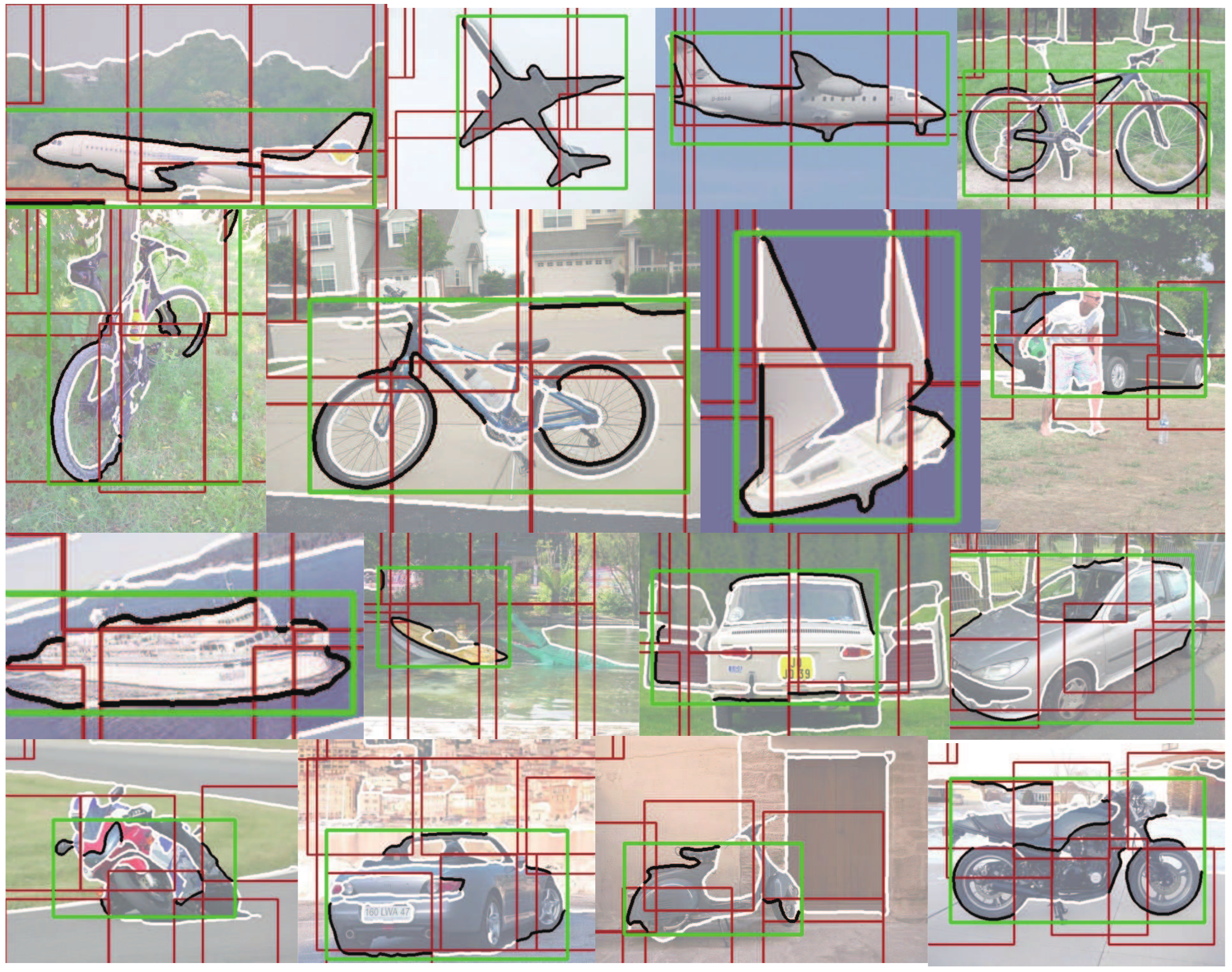,width=0.6\textwidth}
\caption{A few typical object shape detections generated by our approach on the SYSU-Shape dataset. The localized contours are highlighted in black, and the green boxes and red boxes indicate detected shapes and their parts, respectively. }
\label{fig:SYSU_detection}
\end{figure*}

We conduct the experiments on a workstation with Core Duo 3.0 GHZ CPU and 16GB memory. On average, it takes $4 \sim 8$ hours to train a shape model, depending on the numbers of training examples, and the time cost for detection on an image is around $1 \sim 2$ minutes.

{\bf Experiment I.}  We first conduct the experiment on the SYSU-Shape database, which is collected from the Internet and other vision databases. There are $5$ categories, i.e. airplanes, boats, cars, motorbikes, and bicycles, and each category contains $200\sim 500$ images. The shape contours are carefully labeled by a professional team using the LabelMe toolkit~\cite{Labelme}. It is worth mentioning that  each image has at least but not limited to one object of a given category. For each category, half of the images are randomly selected as positive samples and the rest for testing. The images from the other categories are randomly split into two halves as negative samples for training and testing.

%$200\sim 250 $ images are randomly selected from the other categories as negative samples.

\begin{figure*}[!htb]
\centering
\epsfig{figure=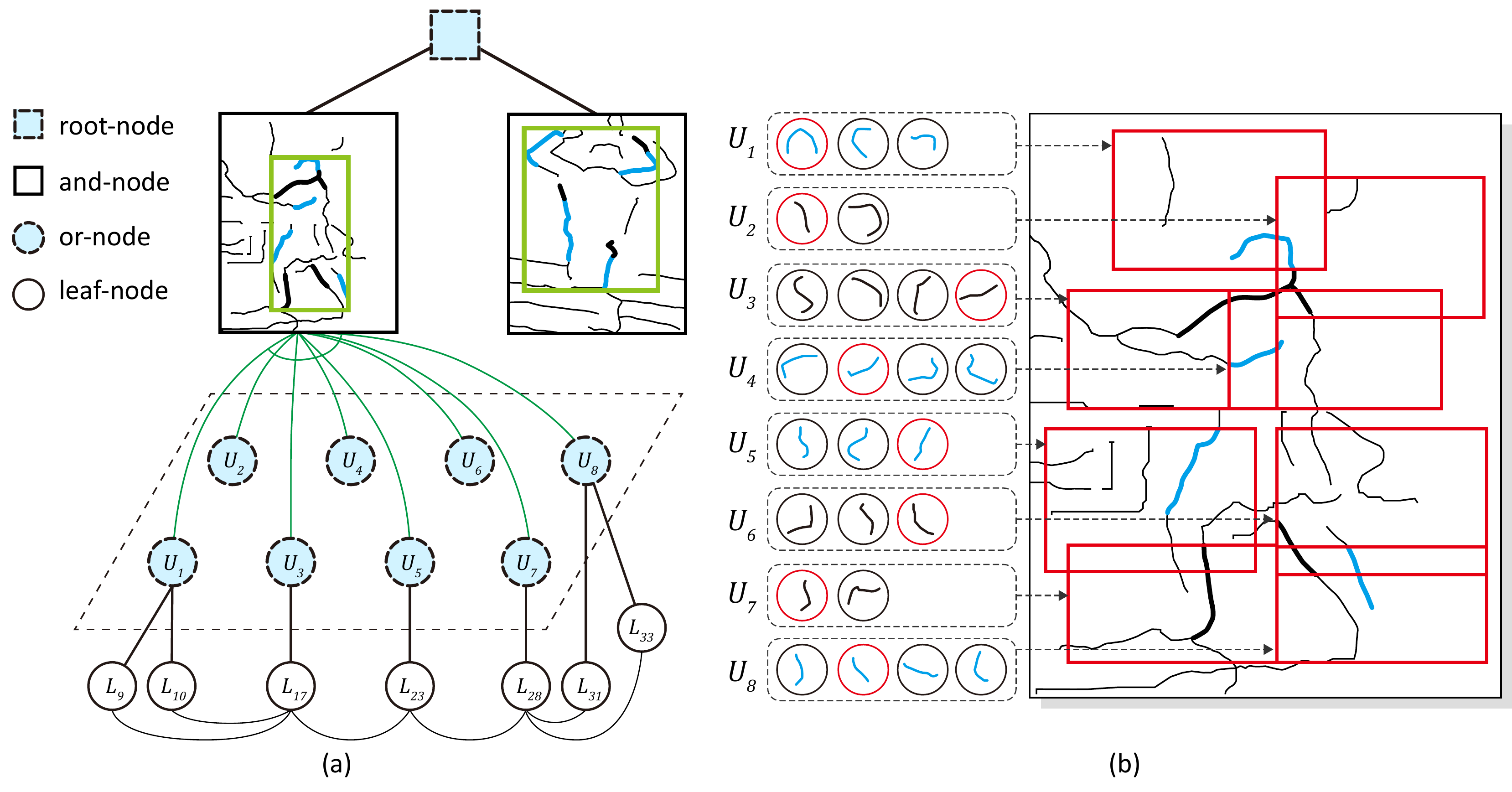,width=0.6\textwidth}
\caption{The trained And-Or graph model with the UIUC-People dataset. (a) Visualizes the model of $4$ layers. (b) Exhibits leaf-nodes associated with or-nodes, $U_1, \ldots, U_8$. A real detection case with the activated leaf-nodes are highlighted in red. }
\label{fig:exp1}
\end{figure*}

\begin{figure*}[!htb]
\centering
\epsfig{figure=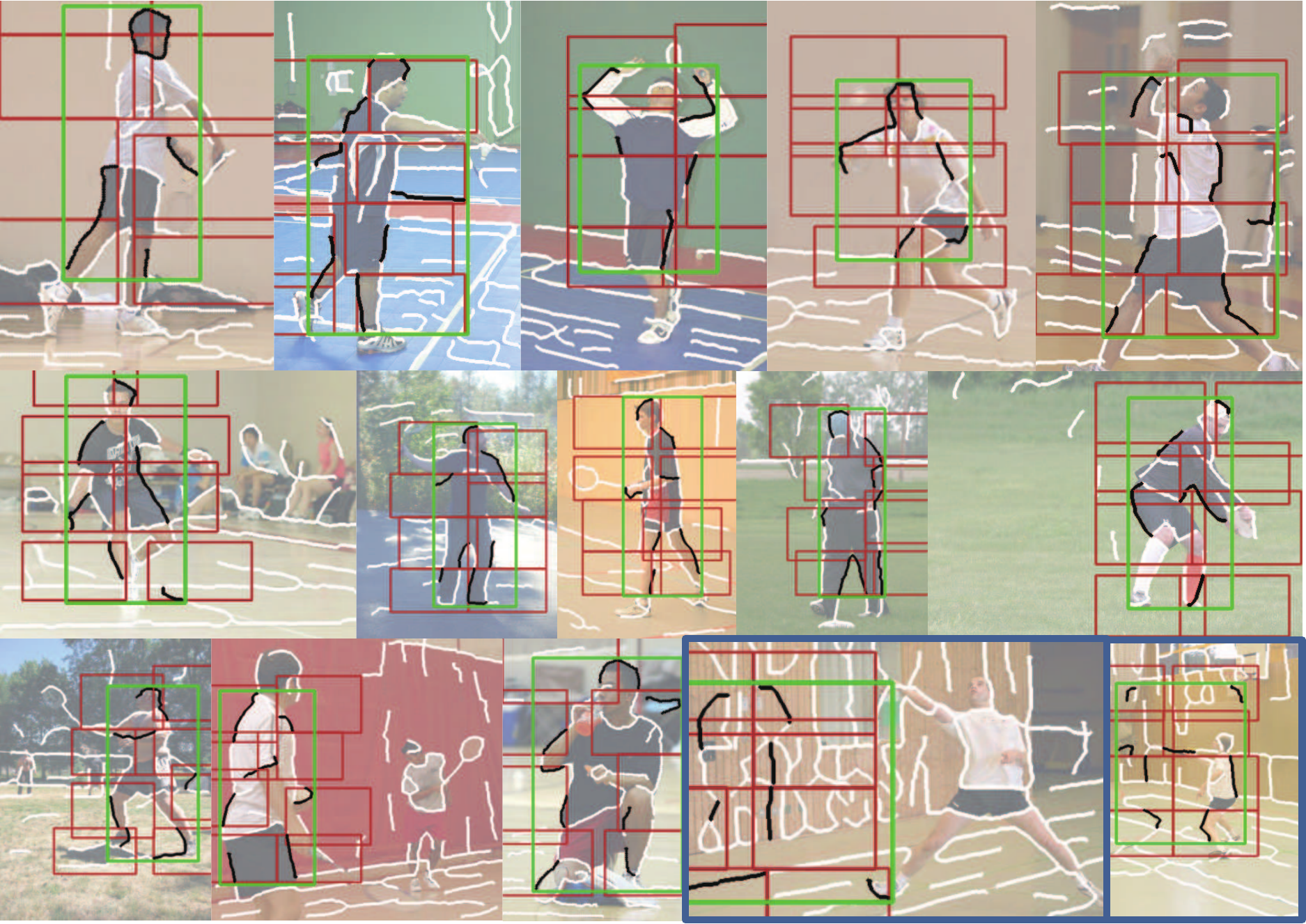,width=0.6\textwidth}
\caption{A few typical object shape detections generated by our method on the UIUC-People database~\cite{UIUCHuman}. The localized contours are highlighted in black, and the green boxes and red boxes indicate detected peoples and parts, respectively. Two failure detections are indicated by the blue boxes.}\label{fig:Results_UIUC}
\end{figure*}

For comparison, we apply the well acknowledged deformable part-based models (DPMs)~\cite{LatentSVM} on this database, where we modify the released code by replacing the input feature with our shape descriptor, and keep the other settings. In this implementation, $3$ DPMs are merged into a mixture, which accounts for different object views. Moreover, we simplify the model into a 3-layer configuration by setting $m = 1$, and test its performances. Figure ~\ref{fig:PR_SYSU_shape} shows the Precision-Recall curves for all $5$ categories, and the Average Precision values are reported in Table~\ref{table:SYSU_shape}. Our complete model achieves the best mean AP and the best APs for all $5$ categories, and the results clearly demonstrate the benefit of using the layered And-Or structures. Several representative detection results are exhibited in Figure~\ref{fig:SYSU_detection}.

%\begin{center}
%\begin{table}[t]	
%		\centering
%
%\begin{tabular}{ c c c c c c c }
%  \hline
%    & \textbf{VOC}  & \textbf{Caltech 101} &  \textbf{Train obj} & \textbf{Test obj} \\
%     \textbf{Aeroplane} & 238 & 262   & &    \\
%     \textbf{Boat} & 257 &  245 &   &   \\
%     \textbf{Car}  & 500 & 0 &   &\\
%     \textbf{Motorbike} & 294 & 206 &   &   \\
%     \textbf{Chair} & 500 &  0 & &   \\
%     \textbf{Bike} & 500 & 0  & &  \\
%  \hline
%\end{tabular}
%		\vspace{3mm}
%		\caption{Statistics of the XX dataset.}\label{table:our-dataset}
%\end{table}
%\end{center}

\begin{figure*}[!htb]
\centering
\epsfig{figure=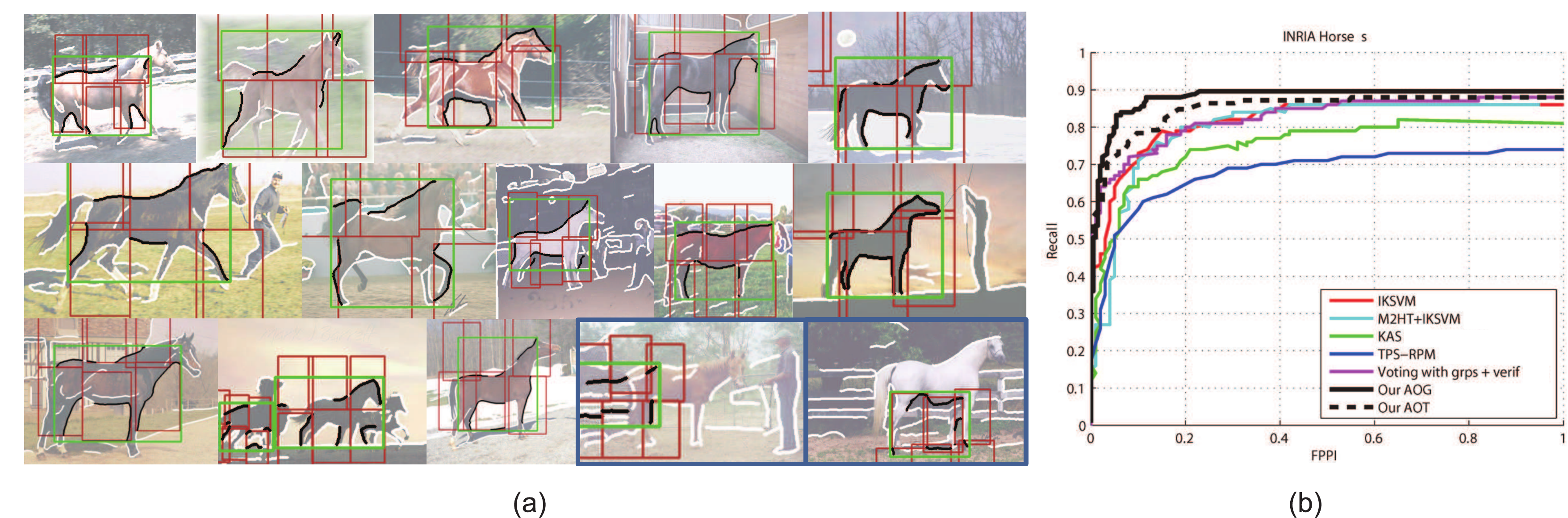,width=1.0\textwidth}
\caption{Results on the INRIA-Horse database. (a) shows several detected shapes  by our method, where the localized contours are highlighted in black, and two failure detections are indicated by the blue boxes. (b) shows the quantitative results with the recall-FPPI measurement. }\label{fig:Horse_fppi}
\end{figure*}

\begin{center}
\begin{table}[t]	
		\centering\begin{tabular*}{0.3\textwidth}{  p{3cm}p{1.1cm} l c } \toprule
         Method        & Accuracy \\
        \hline
         AOG model                                    & \textbf{0.708} \\
%        AOT                                     & 0.670 \\
        Wang et al.~\cite{HierarchicalPoslets}     & 0.668 \\
        Andriluka et al.~\cite{HumanExp1}                      & 0.506  \\
        Felz et al.~\cite{LatentSVM}                      & 0.486  \\
        Bourdev et al.~\cite{HumanExp2}                      & 0.458  \\ \bottomrule
        \end{tabular*}
		\vspace{3mm}
		\caption{Comparisons of detection accuracies on the UIUC-People dataset.}\label{table:UIUC}
\end{table}
\end{center}

{\bf Experiment II.} The UIUC-People dataset contains $593$ images ($346$ for training, $247$ for testing) that are very challenging due to large shape variations caused by different views and human poses. Most of the images contain people playing badminton. The existing methods~\cite{HierarchicalPoslets,HumanExp2}  that are tested on this dataset usually rely on rich appearance-based image features and/or manually labeled prior models. To the best of our knowledge, this work is the first shape-based detector to achieve comparable performances on this dataset. Figure~\ref{fig:exp1}(a) shows the trained And-Or model (AOG), which includes $2$ and-nodes and $8$ or-nodes, and each or-node is associated with $2 \sim 4$ leaf-nodes.  Since most of the images contain one person, we only consider the detection with the highest score on an image for all of the methods. Table~\ref{table:UIUC}  reports the quantitative detection accuracies generated by our method and the competing approaches~\cite{HierarchicalPoslets,HumanExp1,HumanExp2,LatentSVM}. The results (except ours) come from~\cite{HierarchicalPoslets}. A number of representative detection results are presented in Figure~\ref{fig:Results_UIUC}, where the localized contours are highlighted in black, and the green boxes and red boxes indicate detected human and parts, respectively. We also present several inaccurate detections indicated by the blue boxes in Figure~\ref{fig:Results_UIUC}. There are two main reasons for the failure cases: (i) False positives are sometimes created by the background contours segments that appear like the objects-of-interest very much. (ii) The object contours are insufficiently discriminative for recognition, particularly with unconventional object poses and views.

%\begin{figure}[!htb]
%\centering
%\epsfig{figure=exp_final.pdf,width=\textwidth}
%\caption{(a)Experimental results with the recall-FPPI measurement on the INRIA-Horse database. (b),(c) and (d) shows a few object shape detections by applying our method on the three datasets, and the false positives are annotated by blue frames. More results are presented in the supplemental material.}\label{fig:Horse_fppi}
%\end{figure}

{\bf Experiment III.} The INRIA Horse dataset comprise $170$ horse images and $170$ images without horses. The challenges of this dataset arise from background clutter and large deformations, and some of the images contain more than one horse. Following the common experiment setting, we use $50$ positive examples and $80$ negative examples for training and the remaining $210$ images for testing.

Some typical shape detection results on the INRIA Horse dataset are shown in Figure~\ref{fig:Horse_fppi}(a). Compared with existing approaches, we use the recall-FPPI (false positive per image) curves for evaluation, as Figure~\ref{fig:Horse_fppi}(b) reports. It is shown that our approach (denoted as AOG) substantially outperforms the competing methods. Our model achieves detection rates of $89.6\%$ at $1.0$ FPPI; in contrast, the results of competing methods are: $87.3\%$ in ~\cite{VotingECCV2010}, $85.27\%$ in~\cite{MalikCVPR2009}, $80.77\%$ in~\cite{PAS}, and $73.75\%$ in~\cite{FerrariIJCV09}.

{\bf Empirical analysis.} For further evaluation, we present two empirical analysis under different model settings as follows.

(I) We validate the benefit of the contextual collaborative edges. Our model can be further transferred into a tree structure by removing the interactions, which is denoted as ``And-Or Tree (AOT)''. On the UIUC-People dataset, the detection accuracy of the AOT model is $0.69$, which is lower than the complete form of our model, but it is also comparable to the state-of-arts. On the INRIA-Horse dataset, we also present the results yielded by the AOT model in Figure~\ref{fig:Horse_fppi}(b). Based on these results, we can observe that the collaborative edges effectively boost the detection against disturbing surrounding clutter and occlusions.

\begin{figure}[!ht]
\centering
\epsfig{figure=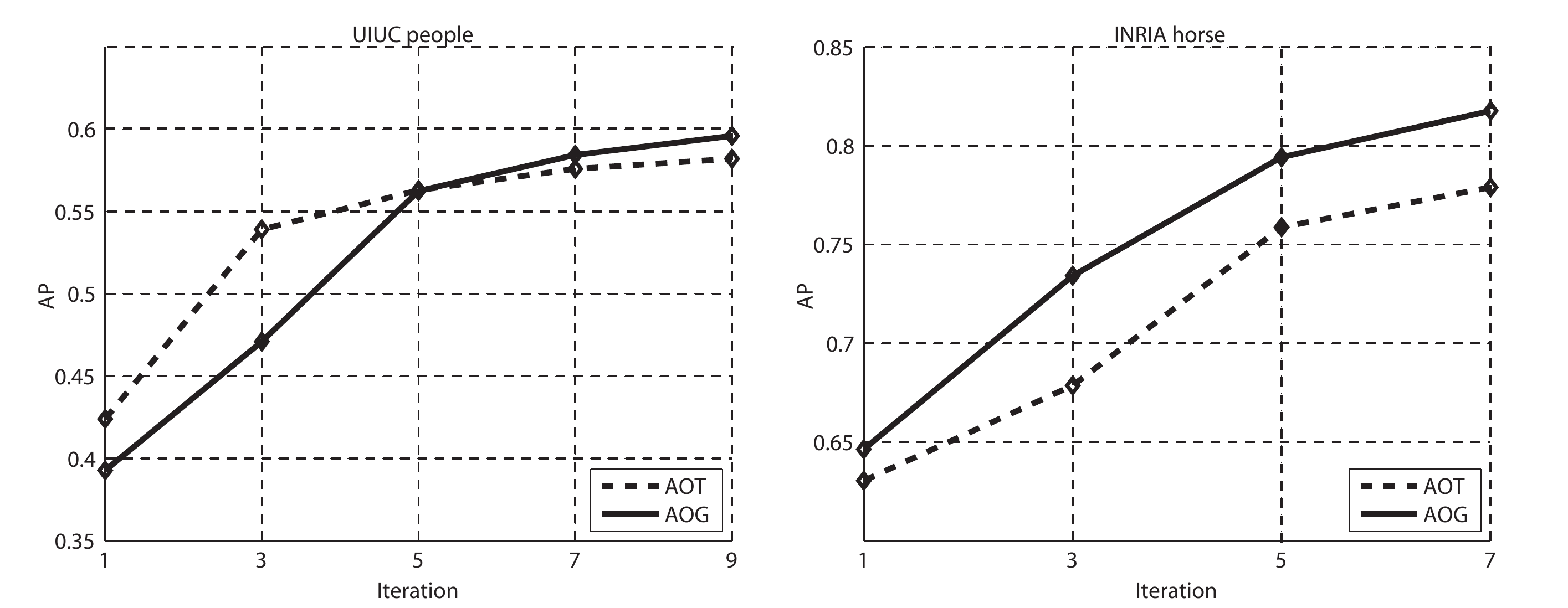,width=0.5\textwidth}
\caption{Model capabilities during the iterative training. We plot the average-precision (AP) with the increasing iterations: the intermediate performances of our models in the iteration steps. We conduct the experiments on the UIUC-People dataset (on the left) and INRIA Horse dataset (on the right). The results of disabling the collaborative edges are also reported.}\label{fig:AP-iteration}
\end{figure}

%\begin{figure*}[!htb]
%\centering
%\epsfig{figure=AOT_AOG_cm.pdf,width=0.8\textwidth}
%\caption{Comparisons of shape detection with the AOG model (in per row) and the AOT model (in the lower row), respectively.}\label{fig:AoT_AoG}
%\end{figure*}

(II) To analyze the model capacity during the iterative training, we output the intermediate performance measures of our models in the iteration steps.

We execute the experiments on the UIUC-People and the INRIA-Horse databases. The quantitative results represented by average precisions (APs) are visualized in Figure~\ref{fig:AP-iteration}. We also report the results generated by the models without collaborative edges, i.e. AOT models. We observe that the discriminative capabilities of our model increase proportinately with the iterations, and converge after a few rounds.

\section{Conclusion and Future Work}\label{sec:conclusion}

In this paper, we have introduced, first, a hierarchical and reconfigurable object shape model in the form of an And-Or graph representation. Second, an efficient inference algorithm for shape detection with the proposed model. Third, a principled learning method that iteratively determine the model structures while optimizing multi-layer parameters.  We demonstrated the practical applicability of our approach by effectively detecting and localizing object shapes from cluttered edge maps. Our model effectively captured large shape variations in deformation for different views and poses. Experiments were implemented on several very challenging databases, (e.g. SYSU-Shapes, UIUC-People, and INRIA-Horse), and our model outperformed other current state-of-the-art approaches.

There are several directions in which we intend to extend this work. The first is to complement our contour-based features with rich appearance information, thereby adapting our model to more general object recognition. The second is to generalize our model in the context of multiclass recognition and investigate part-based structure sharing among classes. For example, the feet of horse and sheep have similar appearances, and thus can be detected by the same local classifier, that is, we can make local classifiers (i.e. the leaf-nodes in our model) shared across categories. Model sharing will keep the model compact while representing multiple categories. Moreover, the inference algorithm will be revised accordingly, to deal with a large number of candidate compositions.

\bibliographystyle{ieee}

\begin{thebibliography}{\small}



\bibitem{HumanExp1}
M. Andriluka, S. Roth, and B. Schiele, Pictorial structures revisited: People detection and articulated pose estimation, In {\em  Proc. of IEEE Conference on Computer Vision and Pattern Recognition (CVPR)}, 2009.

%\bibitem{BaiShapeBand}
%X. Bai, Q. Li, L. J. Latecki, W. Liu, and Z. Tu, Shape band: A deformable object detection approach, In {\em Proc. of IEEE Conference on Computer Vision and Pattern Recognition (CVPR)}, 2009.

\bibitem{BaiTransduction}
X. Bai, X. Yang, L. J. Latecki, W. Liu, and Z. Tu, Learning Context Sensitive Shape Similarity by Graph Transduction, {\em IEEE Trans. on Pattern Analysis and Machine Intelligence}, 32(5): 861-874, 2010.

\bibitem{ShapeContext}
S. Belongie, J. Malik, and J. Puzicha, Shape Matching and Object Recognition using Shape Contexts, {\em IEEE Transactions on Pattern Analysis and Machine Intelligence}, 24(1): 705-522, 2002.


\bibitem{HumanExp2}
L. Bourdev, S. Maji, T. Brox, and J. Malik, Detecting people using mutually consistent poselet activations, In {\em Proc. of European Conference on Computer Vision (ECCV)}, 2010.

\bibitem{DesaiIJCV}
C. Desai, D. Ramanan, and C. C. Fowlkes, Discriminative Models for Multi-Class Object Layout, {\em International Journal of Computer Vision}, 2011.

%\bibitem{DrawingShape2006}
%G. Elidan, G. Heitz, and D. Koller, Learning object shape: From drawings to images, In {\em Proc. of IEEE Conference on Computer Vision and Pattern Recognition (CVPR)}, 2006.

\bibitem{FerrariIJCV09}
V. Ferrari, F. Jurie, and C. Schmid, From Images to Shape Models for Object Detection, {\em International Journal of Computer Vision}, 87(3): 284-303, 2010.

\bibitem{LatentSVM}
P. Felzenszwalb, R. Girshick, D. McAllester, and D. Ramanan, Object Detection with Discriminatively Trained Part-based Models, {\em IEEE Transactions on Pattern Analysis and Machine Intelligence}, 32(9): 1627-1645, 2010.


\bibitem{ShapeTree}
P. Felzenszwalb, and J. D. Schwartz, Hierarchical Matching of Deformable Shapes, In {\em Proc. of IEEE Conference on Computer Vision and Pattern Recognition (CVPR)}, 2007.

\bibitem{Pictorial}
Felzenszwalb, Pedro F., and Daniel P. Huttenlocher. Pictorial structures for object recognition, {\em International Journal of Computer Vision}, 61(1)55-79,  2005.


%\bibitem{FerrariCVPR07}
%V. Ferrari, F. Jurie , and C. Schmid, Accurate Object Detection with Deformable Shape Models Learnt from Images, In {\em CVPR}, 2007.
%




\bibitem{PAS}
V. Ferrari, L. Fevrier, F. Jerie, and C. Schmid, Groups of Adjacent Contour Segments for Object Detection, {\em IEEE Transactions on Pattern Analysis and Machine Intelligence}, 30(1): 36-51, 2008.

\bibitem{CuttingPlane}
V. Franc and S. Sonnenburg, Optimized Cutting Plane Algorithm for Support Vector Machines, In {\em Proc. of International Conference on Machine Learning (ICML)}, 2008.


\bibitem{INRIAHorse}
F. Jurie and C. Schmid, Scale-invariant Shape Features for Recognition of Object Categories, In {\em Proc. of IEEE Conference on Computer Vision and Pattern Recognition (CVPR)}, 2004.

%\bibitem{AlanShapeIJCV2011}
%I. Kokkinos and A. Yuille, Inference and Learning with Hierarchical Shape Models, {\em International Journal of Computer Vision}, 93: 201-225, 2011.

%\bibitem{DimRedPCA}
%N. Kambhatla and T. K. Leen, Dimension Reduction by Local Principal Component Analysis, {\em Neural Computation}, 9(7): 1493-1516, 1997.

\bibitem{ShapeParsing}
I. Kokkinos, and A. Yuille, Inference and Learning with Hierarchical Shape Models, {\em International Journal of Computer Vision}, 93: 201-225, 2011.

\bibitem{LinGraphMatch}
L. Lin, X. Liu, and S.C. Zhu, Layered Graph Matching with Composite Cluster Sampling, {\em IEEE Transactions on Pattern Analysis and Machine Intelligence}, 32(8): 1426-1442, 2010.

\bibitem{LinGrammar}
L. Lin, T. Wu, J. Porway, and Z. Xu, A Stochastic Graph Grammar for Compositional Object Representation and Recognition, {\em Pattern Recognition}, 42(7): 1297-1307, 2009.

\bibitem{LinICCV07}
L. Lin, S. Peng, J. Porway, S.C. Zhu, and Y. Wang, An Empirical Study of
Object Category Recognition: Sequential Testing with Generalized Samples, In {\em Proc. of International Conference on Computer Vision (ICCV)}, 2007.

\bibitem{LinAndOrTree}
L. Lin, X. Wang, W. Yang, and J. Lai, Learning Contour-Fragment-based Shape Model with And-Or Tree Representation, In {\em Proc. of IEEE Conference on Computer Vision and Pattern Recognition (CVPR)}, 2012.

\bibitem{InnerDis}
H. Ling, and D.W. Jacobs, Shape Classification Using the Inner-Distance, {\em IEEE Transactions on Pattern Analysis and Machine Intelligence}, 29(2):286-299, 2007.

\bibitem{LinECCV2010}
P. Luo, L. Lin, and H. Chao, Learning Shape Detector by Quantizing Curve Segments with Multiple Distance Metrics, In {\em Proc. of European Conference on Computer Vision (ECCV)}, 2010.


\bibitem{ShapeGroup}
C. Lu, L. J. Latecki, N. Adluru, X. Yang, and H. Ling, Shape Guided Contour Grouping with Particle Filters, In {\em Proc. of International Conference on Computer Vision (ICCV)}, 2009.


\bibitem{LateckiCVPR2011}
T. Ma and L. J. Latecki, From Partial Shape Matching through Local Deformation to Robust Global Shape Similarity for Object Detection, In {\em Proc. of IEEE Conference on Computer Vision and Pattern Recognition (CVPR)}, 2011.


\bibitem{MalikCVPR2009}
S. Maji and J. Malik, Object Detection using a Max-Margin Hough Transform, In {\em Proc. of IEEE Conference on Computer Vision and Pattern Recognition (CVPR)}, 2009.


\bibitem{PbDetector}
D. Martin, C. Fowlkes, and J. Malik, Learning to detect natural image boundaries using local brightness, color, and texture cues, {\em IEEE Transactions on Pattern Analysis and Machine Intelligence}, 26(5): 530-549, 2004.


\bibitem{ZissermanECCV2006}
A. Opelt, A. Pinz, and A. Zisserman, A Boundary-Fragment-Model for Object Detection, {\em Proceedings of the European Conference on Computer Vision (ECCV)}, 2006.

%\bibitem{SMO}
%J. C. Platt, Using analytic qp and sparseness to speed training of support vector machines, In {\em Proc. Advances in Neural Information Processing Systems (NIPS)}, 1998.




\bibitem{PartialMatchingECCV2010}
H. Riemenschneider, M. Donoser, and H. Bischof, Using Partial Edge Contour Matches for Efficient Object Category Localization, In {\em Proc. of European Conference on Computer Vision (ECCV)}, 2009.

\bibitem{Labelme}
B. C. Russell, A. Torralba, K. P. Murphy, W. T. Freeman, LabelMe: a database and web-based tool for image annotation.\emph{ International Journal of Computer Vision}, pages 157-173, Volume 77, Numbers 1-3, May, 2008.

\bibitem{HierachicalCVPR2009}
P. Schnitzspan, M. Fritz, S. Roth, and B. Schiele, Discriminative structure learning of hierarchical representations for object detection, In {\em Proc. of IEEE Conference on Computer Vision and Pattern Recognition (CVPR)}, 2009.

\bibitem{ShottonPAMI08}
J. Shotton, A. Blake, and R. Cipolla, Multi-Scale Categorical Object Recognition Using Contour Fragments, {\em IEEE Transactions on Pattern Analysis and Machine Intelligence}, 30(7): 1270-1281, 2008.


\bibitem{EventGrammar}
Z. Si, M. Pei, Z.Y. Yao, and S.C. Zhu, Unsupervised Learning of Event And-Or Grammar and Semantics from Video, In {\em  Proc. of International Conference on Computer Vision (ICCV)}, 2011.

\bibitem{AOTemplate}
Z. Si, and S. C. Zhu, Learning And-Or Templates for Object Modeling and Recognition, {\em IEEE Transactions on Pattern Analysis and Machine Intelligence}, 2013.


\bibitem{ContextSVMCVPR2010}
Z. Song, Q. Chen, Z. Huang, Y. Hua, and S. Yan, Contextualizing Object Detection and Classification, In {\em Proc. of IEEE Conference on Computer Vision and Pattern Recognition (CVPR)}, 2010.


\bibitem{ShiShapeCVPR2010}
P. Srinivasan, Q. Zhu, and J. Shi, Many-to-one Contour Matching for Describing and Discriminating Object Shape, In {\em Proc. of IEEE Conference on Computer Vision and Pattern Recognition (CVPR)}, 2010.


\bibitem{UIUCHuman}
D. Tran and D. Forsyth, Improved human parsing with a full relational model, In {\em Proc. of European Conference on Computer Vision (ECCV)}, 2010.

\bibitem{TuShape}
Z. Tu, S. Zheng, and A. Yuille, Shape Matching and Registration by Data-driven EM, {\em Computer Vision and Image Understanding}, 109(3): 290-304, 2008.


\bibitem{ActiveBasis}
Y. Wu, Z. Si, H. Gong, and S.C. Zhu, Learning Active Basis Model for Object Detection and Recognition, {\em International Journal of Computer Visison}, 90(2): 198-235, 2010.


\bibitem{ShapeNIPS2012}
X. Wang, and L. Lin, Dynamical And-Or Graph Learning for Object Shape Modeling and Detection, In {\em Proc. of Advances in Neural Information Processing Systems (NIPS)}, 2012.

\bibitem{HierarchicalPoslets}
Y. Wang, D. Tran, and Z. Liao, Learning Hierarchical Poselets for Human Parsing, In {\em  Proc. of IEEE Conference on Computer Vision and Pattern Recognition (CVPR)}, 2011.


\bibitem{PsychologyContour}
J. De Winter and J. Wagemans. Contour-based object identification and segmentation: stimuli, norms and data, and software tools, {\em Behavior Research Methods,  Instruments, and Computers}, 36(4): 604-624, 2004.


\bibitem{ConFlexibility}
C. Xu, J. Liu, and X. Tang, 2D Shape Matching by Contour Flexibility, {\em IEEE Transactions on Pattern Analysis and Machine Intelligence}, 31(1): 180-186, 2009.


\bibitem{LateckiECCV2010}
X. Yang and L. J. Latecki, Weakly Supervised Shape Based Object Detection with Particle Filter, In {\em Proc. of European Conference on Computer Vision (ECCV)}, 2010.


\bibitem{VotingECCV2010}
P. Yarlagadda, A. Monroy and B. Ommer, Voting by Grouping Dependent Parts, In {\em Proc. of European Conference on Computer Vision (ECCV)}, 2010.


\bibitem{SVMICML2009}
C.-N. J. Yu, and T. Joachims, Learning structural svms with latent variables, In {\em Proc. of International Conference on Machine Learning (ICML)}, 2009.


\bibitem{CCCP}
A.L. Yuille, and A. Rangarajan, The Concave-Convex Procedure (CCCP). {\em Neural Computation}, 15(4): 915-936, 2003.



\bibitem{LeoCCCP}
L. Zhu, Y. Chen, A. Yuille, and W. Freeman, Latent Hierarchical Structural Learning for Object Detection, In {\em Proc. of IEEE Conference on Computer Vision and Pattern Recognition (CVPR)}, 2010.


\bibitem{LeoAOG}
L. Zhu, Y. Chen, Y. Lu, C. Lin, and A. Yuille, Max Margin AND/OR Graph learning for parsing the human body, In {\em Proc. of IEEE Conference on Computer Vision and Pattern Recognition (CVPR)}, 2008.



\bibitem{AOGgrammar}
S.C. Zhu and D. Mumford, A stochastic grammar of images, {\em Foundations and Trends in Computer Graphics and Vision}, 2(4): 259-362, 2006.

\bibitem{ShiShapeECCV2008}
Q. Zhu, L. Wang, Y. Wu, and J. Shi, Contour Context Selection for Object Detection: A Set-to-Set Contour Matching Approach, In {\em Proc. of European Conference on Computer Vision (ECCV)}, 2008.

\vspace{-10mm}


\end{thebibliography}

\begin{IEEEbiography}[{\includegraphics[width=1.0in,clip,keepaspectratio]{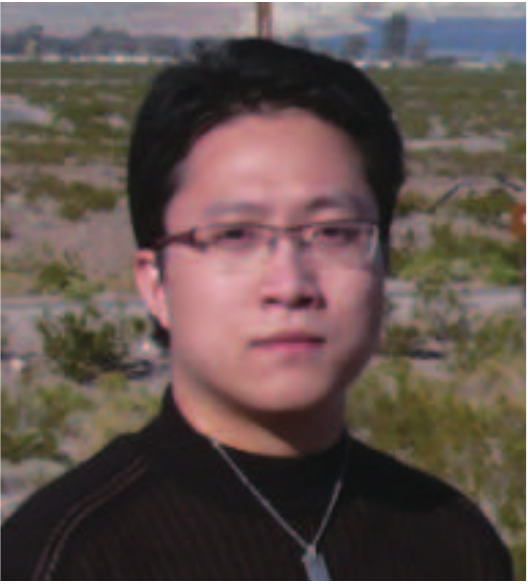}}]{Liang Lin}
is a full Professor with the School of Advanced Computing, Sun Yat-Sen University (SYSU), China. He received the B.S. and Ph.D. degrees from the Beijing Institute of Technology (BIT), Beijing, China, in 1999 and 2008, respectively. From 2006 to 2007, he was a joint Ph.D. student with the Department of Statistics, University of California, Los Angeles (UCLA). He was a Post-Doctoral Research Fellow with the Center for Vision, Cognition, Learning, and Art of UCLA. His research focuses on new models, algorithms and systems for intelligent processing and understanding of visual data such as images and videos. He has published more than 20 papers in highly ranked academic journals and more than 30 papers in top tier conferences CVPR, ICCV, ECCV, ACM MM and NIPS. He was supported by several promotive programs or funds for his works, such as "Program for New Century Excellent Talents" of Ministry of Education (China) in 2012 and Guangdong Distinguished Young Scholar Fund in 2013. He was a recipient of Best Paper Runners-Up Award in ACM NPAR 2010, Google Faculty Award in 2012, and Best Student Paper Award in IEEE ICME 2014. 
\end{IEEEbiography}
\vspace{-10mm}
\begin{IEEEbiography}[{\includegraphics[width=1.0in,clip,keepaspectratio]{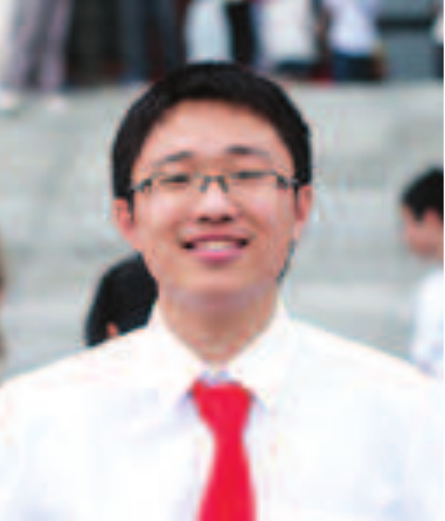}}]{Xiaolong  Wang} 
received the BS degree from the School of Information, South China Agricultural University in 2011, and the MS degree in Computer Science from Sun Yat-sen University in 2014. His research interest include computer vision and machine learning. 
\end{IEEEbiography}
\vspace{-10mm}
\begin{IEEEbiography}[{\includegraphics[width=1.0in,clip,keepaspectratio]{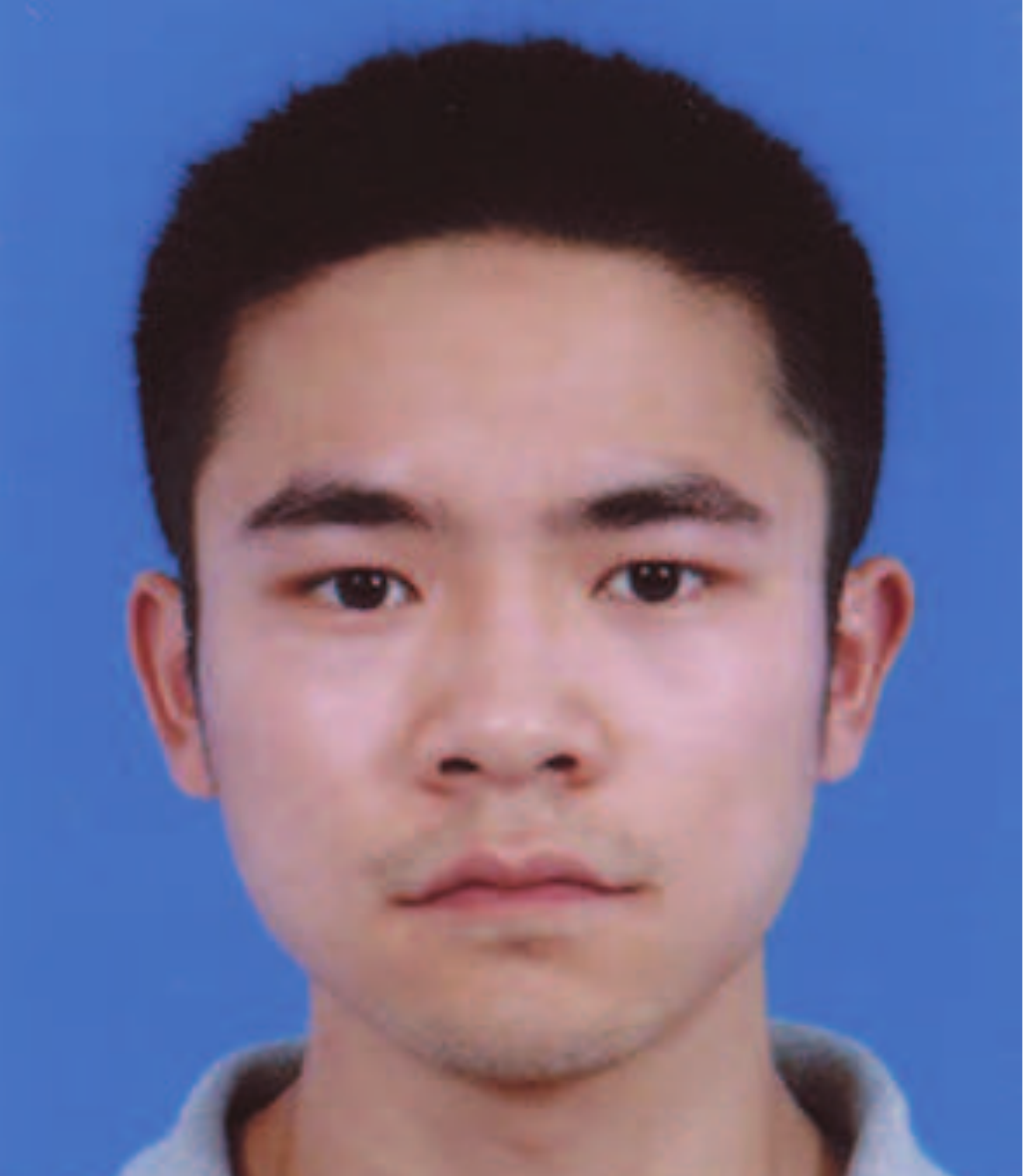}}]{Wei Yang} received the BS degree in Software Engineering from Sun Yat-sen University in 2011, and the MS degree in Computer Science from Sun Yat-sen University in 2014. His research interest include computer vision and machine learning. 
\end{IEEEbiography}
\vspace{-10mm}

\begin{IEEEbiography}[{\includegraphics[width=1in,height=1.25in,clip,keepaspectratio]{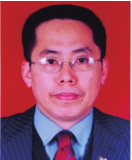}}]{Jianhuang Lai}received his M.Sc. degree in applied mathematics in 1989 and his Ph.D. in mathematics in 1999 from SUN YAT-SEN University, China. He joined Sun Yat-sen University in 1989 as an Assistant Professor, where currently, he is a Professor with the Department of Automation of School of Information Science and Technology and dean of School of Information Science and Technology. His current research interests are in the areas of digital image processing, pattern recognition, multimedia communication, wavelet and its applications. 
\end{IEEEbiography}

\end{document}